\documentclass[a4paper,10pt]{article}

\usepackage[a4paper, total={6in, 9.5in}]{geometry}

\usepackage[utf8]{inputenc} 
\usepackage[T1]{fontenc}    
\usepackage{hyperref}       
\usepackage{url}            
\usepackage{booktabs}       
\usepackage{amsfonts}       
\usepackage{nicefrac}       
\usepackage{microtype}      
\usepackage{xcolor}         
\usepackage{ifthen}
\usepackage{dsfont}

\usepackage{graphicx,wrapfig}
\usepackage[export]{adjustbox}
\usepackage{caption}
\usepackage{subcaption}
\usepackage{dsfont}
\usepackage{amsmath}
\usepackage{amssymb}
\usepackage{mathtools}
\usepackage{tcolorbox}
\usepackage{algorithm}
\usepackage{grffile}
\usepackage{algpseudocode}
\usepackage{makecell}
\usepackage{textcomp}
\usepackage{tikz}

\newtheorem{theorem}{Theorem}

\def\diag{\mathrm{diag}}

\def\Id{\mathrm{Id}}

\def\R{\mathbb{R}}

\def\E{\mathbb{E}}

\def\Jc{\mathcal{J}}
\def\Fc{\mathcal{F}}
\def\Bc{\mathcal{B}}
\def\Cc{\mathcal{C}}

\def\f12{\frac{1}{2}}
\def\ind{\mathds{1}}

\def\eqdef{\stackrel{\mathrm{def}}{=}}
\newcommand\red[1]{#1}
\def\paraspace{}
\def\figspace{}

\title{Adaptive scaling of the learning rate by second order automatic differentiation}

\author{Frédéric de Gournay\textsuperscript{1,2} \\ \footnotesize \href{mailto:degourna@insa-toulouse.fr}{degourna@insa-toulouse.fr}
\and Alban Gossard\textsuperscript{1,3} \\ \footnotesize \href{mailto:alban.paul.gossard@gmail.com}{alban.paul.gossard@gmail.com}
\and \footnotesize \textsuperscript{1} Institut de Math\'ematiques de Toulouse; UMR5219; Universit\'e de Toulouse; CNRS
\and \footnotesize \textsuperscript{2} INSA, F-31077 Toulouse, France
\and \footnotesize \textsuperscript{3} UPS, F-31062 Toulouse Cedex 9, France
}

\begin{document}

\maketitle

\begin{abstract}
In the context of the optimization of Deep Neural Networks, we propose to rescale the learning rate using a new technique of automatic differentiation. This technique relies on the computation of the {\em curvature}, a second order information whose computational complexity is in between the computation of the gradient and the one  of the Hessian-vector product.
If $(1C,1M)$ represents respectively the computational time and memory footprint of the gradient method, the new technique increase the overall cost to either $(1.5C,2M)$ or $(2C,1M)$.
This rescaling has the appealing characteristic of having a natural interpretation, it allows the practitioner to choose between exploration of the parameters set and convergence of the algorithm.
The rescaling is adaptive, it depends on the data and on the direction of descent. The numerical experiments highlight the different exploration/convergence regimes. 
\end{abstract}

\section{Introduction}\label{sec:intro}

The optimization of Deep Neural Networks (DNNs) has received tremendous attention over the past years.
Training DNNs amounts to minimize the expectation of non-convex random functions in a high dimensional space $\R^d$.
If $\Jc:\R^d\rightarrow \R$ denotes this expectation, the problem reads
\begin{equation}\label{eq:min:reel}
    \min_{\Theta\in\R^d} \Jc(\Theta),
\end{equation}
with $\Theta$ the parameters. Optimization algorithms compute iteratively $\Theta_k$, an approximation of a minimizer of~\eqref{eq:min:reel} at iteration $k$, by the update rule
\begin{equation}\label{eq:iteration_formula}
    \Theta_{k+1}=\Theta_k-\tau_k \dot \Theta_k,
\end{equation}
where $\tau_k$ is the learning-rate and $\dot \Theta_k$ is the update direction. The choice of $\dot \Theta_k$ encodes the type of algorithm used.
This work focuses on the choice of the learning rate $\tau_k$.

There is a trade-off in the choice of this learning rate.
Indeed high values of $\tau_k$ allows \emph{exploration} of the parameters space and slowly decaying step size ensures \emph{convergence} in accordance to the famous Robbins-Monro algorithm \cite{robbins1951stochastic}.
This decaying condition may be met by defining the step as $\tau_k=\tau_0 k^{-\alpha}$ with $\tau_0$ being the initial step size and $\f12<\alpha<1$ a constant.
The choice of the initial learning rate and its decay are left to practitioners and these hyperparameters have to be tuned manually in order to obtain the best rate of convergence. 
For instance, they can be optimized using a grid-search or by using more intricated strategies \cite{smith2017don}, but in all generality tuning the learning rate and its decay factor is difficult and time consuming. The main issue is that the learning rate has no natural scaling.
The goal of this work is to propose an algorithm that, given a direction $\dot \Theta_k$ finds automatically a scaling of the learning rate.
This rescaling has the following advantages:
\begin{itemize}
    \item The scaling is adaptive, it depends on the data and of the choice of direction $\dot \Theta_k$.
    \item The scaling expresses the convergence vs. exploration trade-off. Multiplying the rescaled learning rate by $1/2$ enforces convergence whereas multiplying it by $1$ allows for exploration of the space of parameters.
\end{itemize}
This rescaling comes at a cost and it has the following disadvantages:
\begin{itemize}
    \item The computational costs and memory footprint of the algorithm goes from $(1C,1M)$ to $(1.5C,2M)$ or $(2C,1M)$.
    \item The rescaling method is only available to algorithms that yield directions of descent, it excludes momentum method and notably Adam-flavored algorithm.
    \item Rescaling  is theoritically limited to functions whose second order derivative exists and does not vanish. This non-vanishing condition can be compensated by $L^2$-regularization.
\end{itemize}

\subsection{Foreword}\label{sec:foreword}
First recall that second order methods for the minimization of a deterministic $\Cc^2$ function $\Theta\mapsto\mathcal J(\Theta)$, with a Hessian that we denote $\nabla^2\Jc$, are based on the second order Taylor expansion at iteration $k$:
\begin{equation}\label{eq:Taylorexp}
    \mathcal J(\Theta_k-\tau_k \dot \Theta_k) \simeq \mathcal J(\Theta_k) - \tau_k \langle\dot \Theta_k, \nabla\mathcal J(\Theta_k)\rangle +\frac{\tau_k^2}{ 2} \langle \nabla^2 \Jc(\Theta_k) \dot \Theta_k,\dot \Theta_k\rangle .
\end{equation}
If the Hessian of $\Jc$ is positive definite, the minimization of the right-hand side leads to the choice
\begin{equation}\label{eq:choicePk}
    \dot \Theta_k=P_k^{-1} \nabla \Jc(\Theta_k) \text{ with }P_k\simeq \nabla^2 \Jc(\Theta_k).
\end{equation}
Once a direction $\dot \Theta_k$ is chosen, another minimization in $\tau_k$ gives
\begin{equation}\label{eq:optimal_step}
    \tau_k = \frac{\langle\dot\Theta_k, \nabla\Jc(\Theta_k)\rangle}{\Vert \dot \Theta_k\Vert^2 c(\Theta_k,\dot\Theta_k)},
\end{equation}
where $c$ is the curvature of the function, and is defined as
\begin{equation}\label{eq:def_curvature}
    c(\Theta_k, \dot\Theta_k) \eqdef \frac{\langle\nabla^2 \Jc(\Theta_k) \dot \Theta_k,\dot \Theta_k\rangle}{\Vert\dot\Theta_k\Vert^2}.
\end{equation}
A second-order driven algorithm can be decomposed in two steps: i) the choice of $P_k$ in~\eqref{eq:choicePk}, and if this choice leads to an update which is a direction of ascent, that is  $\langle \dot \Theta_k ,\nabla \Jc(\Theta_k)\rangle >0$, ii) a choice of $\tau_k$ by an heuristic inspired from~\eqref{eq:def_curvature} and~\eqref{eq:optimal_step}.

In the stochastic setting, we denote as $s\mapsto \Jc_s$ the mapping of the random function.
At iteration $k$, only information on $(\Jc_s)_{s\in \Bc_k}$ can be computed where $(\Bc_k)_k$ is a sequence of mini-batches which are indepently drawn.
If $\E_{s\in\Bc_k}$ is the empirical average over the mini-batch, we define $\Jc_{\Bc_k}= \E_{s\in\Bc_k}[ \Jc_s]$.
Given $\Theta$, the quantity $\Jc(\Theta)$ is deterministic, and $\Jc$ is the expectation of $\Jc_s$ w.r.t. $s$.

\subsection{Related works}

{\bf Choice of $P_k$:} The choice $P_k=\nabla^2 \Jc_{\Bc_k}(\Theta_k)$ in~\eqref{eq:choicePk}, leads to a choice $\tau_k=1$  and to the so-called Newton method. It is possible in theory to compute the Hessian by automatic differentiation if it is sparse~\cite{walther2008computing}, but to our knowledge it has not been implemented yet. In~\cite{martens2010deep}, the authors solve $\dot \Theta_k=\left[\nabla^2 \Jc_{\Bc_k}(\Theta_k)\right]^{-1} \nabla \Jc_{\Bc_k}(\Theta_k)$ by a conjugate gradient method which requires only matrix-vector product which is affordable by automatic differentiation \cite{christianson1992automatic,pearlmutter1994fast}.
This point of view, as well as some variants \cite{vinyals2012krylov,krishnan2017neumann}, suffer from high computational cost per batch and go through less data in a comparable amount of time, leading to slower convergence at the beginning of the optimization.

Another choice is to set $P_k\simeq\nabla^2 \Jc_{\Bc_k}(\Theta_k)$ in~\eqref{eq:choicePk} which is coined as the ``Quasi-Newton'' approach.
These methods directly invert a diagonal, block-diagonal or low rank approximation of the Hessian \cite{becker1988improving,schaul2013no,roux2007topmoumoute,ollivier2015riemannian,martens2015optimizing,yao2020adahessian}.
In most of these works, the Hessian is approximated by $\E[\nabla \Jc_s(\theta_k)\nabla \Jc_s(\theta_k)^T]$, the so-called Fisher-Information matrix, which leads to the natural gradient method \cite{amari1998natural}.
Note also the use of a low-rank approximation of the true Hessian for variance reduction in \cite{gower2018tracking}.

Finally, there is an interpretation of adaptive methods as Quasi-Newton methods.
Amongst the adaptive method, let us cite RMSProp~\cite{tieleman2012divide}, Adam~\cite{kingma2014adam}, Adagrad~\cite{duchi2011adaptive} and Adadelta~\cite{zeiler2012adadelta}.
For all these methods, $P_k$ is as a diagonal preconditioner that reduces the variability of the step size across the different layers. This class of methods can be written
\begin{equation}
    \dot\Theta_{k} = P_k^{-1}m_k, \quad m_k\simeq \nabla\Jc(\Theta_k)\quad \text{ and }\quad P_k\simeq \nabla^2 \Jc(\Theta_k).
\end{equation}
For instance, RMSProp and Adagrad use $m_k=\nabla\Jc_{\Bc_k}(\Theta_k)$ whereas Adam maintains in $m_k$ an exponential moving averaging from the past evaluations of the gradient.
The RMSProp, Adam and Adagrad optimizers build $P_k$ such that $P_k^2$ is a diagonal matrix whoses elements are exponential moving average of the square of the past gradients (see \cite{reddi2019convergence} for example). It is an estimator of the diagonal part of the Fisher-Information matrix.

All these methods can be incorporated in our framework as we consider the choice of $P_k$ as  a preconditioning technique whose step is yet to be found. In a nutshell, if  $P_k$ approximates the Hessian up to an unknown multiplicative factor, our method is able to find this multiplicative factor.

{\bf Barzilai-Borwein:}
The Barzilai-Borwein (BB) class of methods~\cite{barzilai1988two,raydan1997barzilai,dai2002modified,xiao2010notes,biglari2013scaling,li2019new} may be interpreted as methods which aim at estimating the curvature in~\eqref{eq:def_curvature} by numerical differences using past gradient computations.
In the stochastic convex setting, the BB method is introduced in \cite{tan2016barzilai} for the choice $\dot \Theta_k=\nabla \Jc(\Theta_k)$ and also for variance-reducing methods \cite{johnson2013accelerating}.
It has been extended in \cite{ma2018stochastic} to non-convex problems and in \cite{liang2019barzilai} to DNNs.
Due to the variance of the gradient and possibly to a poor estimation of the curvature by numerical differences, these methods allow prescribing a new step at each epoch only.
In \cite{yang2018random,castera2022second}, the step is prescribed at each iteration at the cost of computing two mini-batch gradients per iteration.
Moreover, in \cite{yang2018random} the gradient over all the data needs to be computed at the beginning of each epoch whereas \cite{castera2022second} maintains an exponential moving average to avoid this extra computation.
The downside of \cite{castera2022second} is that they still need to tune the learning rate and its decay factor and that their method has not been tried on other choices than $P_k=\Id$.

Our belief is that approximating by numerical differences in a stochastic setting suffers too much from variance from the data and from the approximation error. Hence we advocate in this study for exact computations of the curvature~\eqref{eq:def_curvature}.

{\bf Automatic differentiation:}
The theory that allows to compute the matrix-vector product of the Hessian with a certain direction is well-studied \cite{walther2008computing,christianson1992automatic,griewank2008evaluating,pearlmutter1994fast} and costs $4$ passes ($2$ forward and backward passes) and $3$ memory footprint, when the computation of the gradient costs $2$ passes ($1$ forward and backward pass) and $1$ memory footprint.
We study the cost of computing the curvature defined in~\eqref{eq:def_curvature}, which to the best of our knowledge, has never been studied.
Our method has a numerical cost that is always lower than the best BB method \cite{castera2022second}.

\subsection{Our contributions}

We propose a change of point of view. While most of the methods presented above use first order information to develop second order algorithms, we use second order information to tune a first order method.
The curvature \eqref{eq:def_curvature} is computed using automatic differentiation in order to estimate the local Lipschitz constant of the gradient and to choose a step accordingly.
Our contribution is threefold:
\begin{itemize}
    \item We propose a method that automatically rescales the learning rate using curvature information in Section~\ref{sec:rescale} and we discuss the heuristics of this method in Section~\ref{sec:analysis_rescaling}. {The rescaling allows the practitioner to choose between three different physical regimes coined as : hyperexploration, exploration/convergence trade-off and hyperconvergence.}
    \item { We study the automatic differentiation of the curvature in Section~\ref{sec:curvcomputation} and its computational costs.}
    \item {Numerical tests are provided in Section~\ref{sec:num} with a discussion on the three different physical regimes introduced in Section~\ref{sec:rescale}.}
\end{itemize}

\section{Rescaling the learning rate}
\subsection{Presentation and guideline for rescaling}\label{sec:rescale}

The second order analysis of Section~\ref{sec:intro} relies on the Taylor expansion
\[\mathcal J(\Theta_k-\tau_k \dot \Theta_k) \simeq \mathcal J(\Theta_k) - \tau_k \langle\dot \Theta_k, \nabla\mathcal J(\Theta_k)\rangle +\frac{\tau_k^2}{ 2} c(\Theta_k,\dot \Theta_k) \Vert \dot \Theta_k\Vert^2,\]
with $c(\Theta_k,\dot \Theta_k)$ given by~\eqref{eq:def_curvature}. This Taylor expansion yields the following algorithm: given $\Theta_k$ and an update direction $\dot \Theta_k$, compute $c(\Theta_k,\dot \Theta_k)$ by~\eqref{eq:def_curvature}, the step $\tau_k$ by~\eqref{eq:optimal_step} and finally update the parameters $\Theta_k$ by~\eqref{eq:iteration_formula}.
The first order analysis is slightly different.
Starting with the second order exact Taylor expansion in integral form:

\[\mathcal J(\Theta_k-\tau_k \dot \Theta_k) = \mathcal J(\Theta_k) + \tau_k \langle\dot \Theta_k, \nabla\mathcal J(\Theta_k)\rangle + \int_{t=0}^{\tau_k} (\tau_k-t) c(\Theta_k-t\dot \Theta_k,\dot\Theta_k) \Vert \dot \Theta_k\Vert^2dt,\]
we introduce the local directional Lipschitz constant of the gradient
\begin{equation}\label{eq:estimateLips}
    L_k=\max_{t\in [0,\tau_k]} |c(\Theta_k-t\dot \Theta_k,\dot\Theta_k)|,
\end{equation}
in order to bound the right-hand side of the Taylor expansion. One obtains
\begin{equation}\label{eq:quadratic_approx}
    \mathcal J(\Theta_k-\tau_k \dot \Theta_k) \le \mathcal J(\Theta_k) - \tau_k \langle\dot \Theta_k, \nabla\mathcal J(\Theta_k)\rangle +\frac{\tau_k^2}{ 2} L_k \Vert \dot \Theta_k\Vert^2.
\end{equation}
Introducing the rescaling $r_k$
\begin{equation}\label{eq:rescaling}
    r_k =\frac{\langle\dot \Theta_k,\nabla\Jc(\Theta_k)\rangle}{\Vert \dot \Theta_k\Vert^2 L_k }
\end{equation}
and writing $\tau_k=r_k\ell$, Equation \eqref{eq:quadratic_approx} turns into
\begin{equation}\label{eq:quadratic_approx2}
    \Jc(\Theta_k-\ell r_k\dot \Theta_k) \le \mathcal J(\Theta_k) +\left(\ell^2 - \ell \right)\frac{L_k}{2}\Vert r_k\dot \Theta_k\Vert^2 \quad \forall \ell.
\end{equation}
Any choice of $\ell$ in $]0,1[$ leads to a decrease of $\Jc$ in~\eqref{eq:quadratic_approx2}. The choice $\ell=\frac 1 2$ allows faster decrease of the right-hand side of \eqref{eq:quadratic_approx2}. We coin the choice $\ell=1$ in \eqref{eq:quadratic_approx2} as the \emph{exploration choice} and the choice $\ell=\frac 1 2$ as the \emph{convergence choice}.
The only difficulty in computing \eqref{eq:rescaling} lies in the computation of $L_k$. Indeed, $L_k$ is a maximum over an unknown interval and, in the stochastic setting, we only estimate the function $\Jc$ and its derivative on a batch $\mathcal B_k$.

We propose to build $\tilde L_k$ an estimator of $L_k$ by the following rules.
\begin{itemize}
\item Replace the maximum over the unknown interval $[0,\tau_k]$ in \eqref{eq:estimateLips} by the value at $t=0$. This is reminiscent of the Newton's method.
\item Perform an exponential moving average on the past computations of $r_k$ in order to average over the data previously seen.
\item Use the maximum of this latter exponential moving average and the current estimate in order to stablize $\tilde L_k$.
\end{itemize} The algorithm reads as follows:
\begin{algorithm}
\caption{Rescaling of the learning rate}
\begin{algorithmic}[1]
\State {\bf Hyperparameters} $\beta_3=0.9$ (exponential moving average).
\State {\bf Initialization} $\hat c_0=0$ 
\State {\bf Input (at each iteration $k$)}: a batch $\Bc_k$, $g_k=\E_{s\in \Bc_k}\left[\nabla \Jc_s(\Theta_k)\right]$ and $\dot \Theta_k$ a direction that verifies $\langle g_k,\dot \Theta_k\rangle >0$.
  \State $c_k=\E_{s\in \Bc_k}\left[\left|\langle\nabla^2 \Jc_s(\Theta_k)\dot \Theta_k,\dot \Theta_k\rangle\right|\right] / \Vert \dot \Theta_k\Vert^2$ \Comment{local curvature} \label{lst:line:absolute value}
  \State $\hat c_k= \beta_3 \hat c_{k-1}+(1-\beta_3)c_k\quad$ and $\quad\tilde c_k= \hat c_{k}/(1-\beta_3^k)$  \Comment{moving average}  \label{lst:line:ema-c}
  \State $\tilde L_k=\max(\tilde c_k , c_k)$  \Comment{stabilization}   \label{lst:line:max-c}
  \State $r_k= \langle\dot \Theta_k,g_k\rangle/\left(2\Vert \dot \Theta_k\Vert^2\tilde L_k\right)$  \Comment{rescaling factor} \label{lst:line:maxr}
  \State {\bf Output (at each iteration $k$)}: $r_k$ a rescaling of the direction $\dot \Theta_k$.
  \State {\bf Usage of rescaling}: The practitioner should use the update rule $\Theta_{k+1}=\Theta_{k} -\ell r_k\dot \Theta_k$, where $\ell$ follows the {\em Rescaling guidelines} (see below).
\end{algorithmic}
\label{alg:rescale}
\end{algorithm}

Note that the curvature $c_k$ is computed with the same batch that the one used to compute $g_k$ and $\dot \Theta_k$.

\paragraph{Rescaling guidelines} Given a descent direction $\dot \Theta_k$, the update rule is given by 
\[\Theta_{k+1}=\Theta_k-\ell r_k \dot\Theta_k,\]
where $\ell$ is the learning-rate that has to be chosen by the practitioner and $r_k$ is the rescaling computed by Algorithm~\ref{alg:rescale}. In the deterministic case, $\ell$ has a physical interpretation:
\begin{itemize}
    \item $1 \ge \ell \ge \frac 1 2$ ({\em Convergence/exploration trade-off}). The choice $\ell=1$ (exploration) is the largest step that keeps the loss function non increasing. The choice $\ell =1/2$ (convergence) ensures the fastest convergence to the closest local minimum. It is advised to start from $\ell=1$ and decrease to $\ell=1/2$ (see Section~\ref{sec:conv_expl_tradeoff}).
    \item $\ell >1$ ({\em Hyperexploration}). The expected behavior is a loss function increase and large variations of the parameters. This mode can be used to escape local basin of attraction in annealing methods (see Section~\ref{sec:hyperexpl}).
    \item $0<\ell<1/2$ ({\em Hyperconvergence}). This mode slows down the convergence. In the stochastic setting, if the practitioner has to resort to setting $\ell <1/2$ in order to obtain convergence, then some stochastic effects are of importance in the optimization procedure (see Section~\ref{sec:hyperconv}).
\end{itemize}
\paraspace

\subsection{Analysis of the rescaling}\label{sec:analysis_rescaling}

Several remarks are necessary to understand the limitations and applications of rescaling.

\paraspace
\paragraph{The algorithm does not converge in the deterministic setting.}
Note that in the deterministic one dimensional case, when $\Jc$ is convex (i.e. the curvature is positive), $\beta_3=0$ and $\ell=1/2$, the algorithm boils down to the Newton method.
It is known that the Newton method may fail to converge, even for strictly convex smooth functions. For example if we choose
\[\Jc(\Theta)=\sqrt{1+\Theta^2},\]
the iterates of the Newton method are given by $\Theta_{k+1}=-\Theta_k^3$, which diverges as soon as $|\Theta_0|>1$.
This problem comes from the fact that the curvature $c(\Theta_k-t\dot \Theta_k,\dot \Theta_k)$ has to be computed for each $t\in[0,\tau_k]$ in order to estimate $L_k$ in~\eqref{eq:estimateLips} but this maximum is estimated by its value at $t=0$.
In this example, $c(\Theta_k,\dot \Theta_k)$ is a bad estimator for $L_k$ as it is too small and the resulting step is too large.

Another issue in DNN is the massive use of piecewise linear activation functions which can make the Hessian vanish and in this case, the rescaled algorithm may diverge.
For example if the loss function is locally linear, then $c_k=0$ in line~\ref{lst:line:absolute value} and if we choose $\beta_3=0$ then $r_k=+\infty$ in line~\ref{lst:line:maxr}.

\paraspace
\paragraph{The algorithm does not converge in the stochastic setting.} Let $X$ be a vector-valued random variable and $\Jc$ is the function
\[\Jc(\Theta)=\frac 1 2 \E\left[\Vert \Theta -X\Vert^2\right],\]
then for any value of $\beta_3$ and for the choice $\ell=\frac 1 2$, the rescaled algorithm yields  the update $\Theta_{k}=\E_{\mathcal B_k}[X]$, when the optimal value is $\Theta^\star=\E[X]$.
The algorithm oscillates around $\Theta^\star$, with oscillations depending on the variance of the gradient. This oscillating stochastic effect is well known and is the basic analysis of SGD. Since the proposed rescaling analysis is performed in a deterministic setting, it is not designed to offer any solution to this problem.

\paraspace
\paragraph{Enforcing convergence by Robbins-Monro conditions}
In order to enforce convergence, we can use the results of \cite{robbins1951stochastic}. It is then sufficient to sow instructions like
\begin{equation}\label{eq:stab:RM}
    \alpha\le \ell r_k k^{\delta}\le \beta,
\end{equation}
with fixed $\alpha,\beta>0$ and $\delta\in]1/2,1[$. This is the choice followed by \cite{castera2022second} for instance.
Note that convergence analysis for curvature-dependent step is, to our knowledge, studied only in \cite{alvarez2004steepest}, for the non-stochastic time-continuous setting.

\paraspace
\paragraph{Fostering convergence with $L_2$ regularization}
In the deterministic case, the non-convergence of the algorithm is caused by vanishing eigenvalues of the Hessian. This issue can be fixed by adding a term $\frac\lambda 2 \Vert \Theta\Vert^2$ to the function $\Theta \mapsto \Jc(\Theta)$, with $\lambda >0$. This method is coined as $L_2$ regularization with parameter $\lambda$. This method shifts the eigenvalues of the Hessian of $\Jc$ by the parameter $\lambda$. Close to the minimum, every eigenvalue of the Hessian is then positive. Although $L_2$ regularization does not guarantee convergence, it promotes it.

\paraspace
\paragraph{Gradient preconditioning}
In case of gradient preconditioning $\dot \Theta_k=P_k^{-1} g_k$ with $P_k\simeq \nabla^2\Jc(\Theta_k)$, the advantage of rescaling is that the practitioner is allowed to approximate the Hessian up to a multiplicative factor. Indeed suppose that instead of providing a good estimate of the Hessian, the practitioner multiplies it at each iteration by an arbitrary factor $\alpha_k\in \mathbb{R}$. In this case, $\dot \Theta_k$ is multiplied by $\alpha_k^{-1}$ but the curvature $c_k$ does not change. This means that $\hat c_k$ is independent of the previous $(\alpha_s)_{s\le k}$. Finally, the rescaling $r_k$ is multiplied by $\alpha_k$. Hence the output of the algorithm $r_k\dot\Theta_k$ is independent of the sequence $(\alpha_s)_{s\le k}$. Therefore, the practitioner does not need to worry about finding the right multiplicative factor, it is accounted for by the rescaling method.

\paraspace
\paragraph{Negativeness of the curvature (line~\ref{lst:line:absolute value})}
The main difference beween a first-order analysis~\eqref{eq:estimateLips} and a second-order~\eqref{eq:optimal_step} lies in handling the case when the curvature is negative. The first-order analysis, which we choose, relies on using absolute value of the curvature, when second-order analysis relies on more intricated methods, see \cite{allen2018natasha,carmon2017convex,liu2017noisy,curtis2019exploiting}.
Note that the absolute value is taken inside the batch average in line~\ref{lst:line:absolute value} and not outside.
Otherwise data in the batch where $\langle\nabla \mathcal J^2(\Theta_k)\dot \Theta_k,\dot \Theta_k\rangle$ is negative could compensate the data where it is positive, leading to a bad estimation of the curvature.

\paraspace
\paragraph{Heuristics in the estimation of $L_k$ (lines~\ref{lst:line:ema-c} and~\ref{lst:line:max-c})}
The estimator of $L_k$ must comply with two antagonist requirements.
The first one is to average the curvature over the different batches to effectively compute the true curvature of $\Jc$.
The second one is to use the local curvature at point $\Theta_k$ and in the direction $\dot \Theta_k$ which requires to forget old iterations.
This advocates the use of an exponential moving average in line~\ref{lst:line:ema-c} with the parameter $\beta_3$.
The maximum in line~\ref{lst:line:max-c} is reminiscent of the construction of AMSGrad \cite{reddi2019convergence} from Adam \cite{kingma2014adam}, and it stabilizes batches where $c_k \gg \tilde c_k$. In order to be consistent with the remark in \emph{Gradient preconditioning}, the averaged quantity is the one which does not depend on the unknown multiplicative factor $\alpha_k$.

\section{Computing the curvature}\label{sec:curvcomputation}
In this section, we focus on the computation of $c(\Theta,\dot \Theta)$ by automatic differentiation and its cost.

\subsection{Main results}

A Neural Network $\mathcal N$ is a directed acyclic graph and at each node of the graph, the data are transformed and fed to the rest of the graph.
The data at the output $x_n$ are then compared to $y$. Since there is no cycle in the graph, there is no mathematical restriction to turn such graph into a list. The set of parameters for layer $s$ is denoted as $\theta_s$, and we denote $\Theta=(\theta_s)_{s=0..n}$ the set of parameters of $\mathcal N$. The action of $\mathcal N$ is expressed by the recurrence:
\begin{equation}\label{eq:forward:co}
    x_{s+1}(\Theta) =\mathcal F_{s}(x_s(\Theta),\theta_s), \quad 0 \le s\le n-1
\end{equation}
where $\mathcal F_{s}$ is the action of the $s^{th}$ layer of $\mathcal N$.
The output $x_n$ is then compared to a target via a loss function $\mathcal F_n$ and we denote $x_{n+1}\in\R$ the result of this loss function.

Let $X(\Theta)=(x_s(\Theta))_{s=0..n+1}$ denote the set of data as it is transformed through the neural network.
The intermediate data $x_s(\Theta)$ (resp. parameter $\theta_s$) are supposed to belong to an Hilbert space $\mathcal H_s$ (resp. $\mathcal G_s$).
We then have for each $0\le s \le n$ 
\[\mathcal F_{s}: \mathcal H_s \times \mathcal G_s \rightarrow \mathcal H_{s+1}\text{ and }\mathcal H_{n+1}=\mathbb R.\]

The gradient of $\Jc$ with respect to $\Theta$ is computed using automatic differentiation.
This requires to define the differentials of $\Fc_s$ with respect to its variables.
Let $\partial_x \mathcal F_s:\mathcal H_{s} \rightarrow \mathcal H_{s+1}$, resp. $\partial_\theta \mathcal F_s : \mathcal G_{s} \rightarrow \mathcal H_{s+1}$, be the differential of $\mathcal F$ at the point $(x_s(\Theta),\theta_s)$ w.r.t. $x$, resp. $\theta$.
Denote $(\partial_x \mathcal F_s)^*:\mathcal H_{s+1} \rightarrow \mathcal H_s$ and $(\partial_\theta \mathcal F_s)^* : \mathcal H_{s+1} \rightarrow \mathcal G_s$ the adjoints of the differentials of $\mathcal F_s$. These adjoints are defined for all $\phi \in \mathcal H_{s+1}$ as the unique linear mapping that verifies:
\begin{align*}
    \langle \partial_x \mathcal F_s^* \phi ,\psi\rangle_{\mathcal H_s} =& \langle \phi ,\partial_x \mathcal F_s\psi \rangle_{\mathcal H_{s+1}} \quad \forall \psi \in \mathcal H_s \\
    \langle \partial_\theta \mathcal F_s^* \phi ,\psi\rangle_{\mathcal G_s} =& \langle \phi ,\partial_\theta \mathcal F_s\psi \rangle_{\mathcal H_{s+1}} \quad \forall \psi \in \mathcal G_s.
\end{align*}
Denote by $\nabla^2\Fc_s$ the second order derivative tensor of $\Fc_s$ at the point $(x_s,\theta_s)$.
The backward of the data $\hat X=(\hat x_s)_{s=1..n+1}$ and the backward-gradient $\hat \Theta=(\hat \Theta_s)_{s=0..n}$ are defined by:
\begin{equation}
    \label{eq:backward}
    \begin{cases}
        \hat x_s =(\partial_x \mathcal F_s)^*\hat x_{s+1} \quad \text{ with }\hat x_{n+1}=1 \\
        \hat \theta_s =(\partial_\theta \mathcal F_s)^*\hat x_{s+1}.
    \end{cases}
\end{equation}

In Algorithm~\ref{alg:backpropwithcurv}, the standard backpropagation algorithm is given as well as the modifications needed to compute the curvature. The proof of this algorithm is given in Section~\ref{sec:proof_algocurv}.

\definecolor{colorcurv}{rgb}{0.5, 0, 0.5}

\begin{algorithm}[H]
    \caption{Backpropagation {\color{colorcurv} with curvature computation}}
    \label{alg:backpropwithcurv}
    \begin{algorithmic}[1]
    \State Compute and store the data $X=(x_s)_s$ with a forward pass~\eqref{eq:forward:co}.
    \State Compute {\color{colorcurv} and store} the backward $\hat X=(\hat x_s)_s$ and $\hat \Theta=(\hat \theta_s)_s$ using~\eqref{eq:backward}.
    \State Then $\nabla \mathcal J(\Theta)=\hat \Theta.$
        \State Choose any direction of update $\dot \Theta=(\dot \theta_s)_s$.
        {\color{colorcurv}
        \State Compute the tangent $\dot X=(\dot x_s)_s$ with the following forward pass:
        \begin{equation}
            \label{eq:directional:derivative}
             \dot x_{s+1}=(\partial_x \mathcal F_s)\dot x_s+(\partial_\theta \mathcal F_s)\dot \theta_s, \quad \dot x_0=0
        \end{equation}
        \State Then $\langle \nabla^2 \mathcal J(\Theta)\dot \Theta,\dot \Theta\rangle = \sum_s \langle  \hat x_{s+1},\nabla^2 \mathcal F_s(\dot x_s,\dot \theta_s)\otimes (\dot x_s,\dot \theta_s) \rangle_{\mathcal H_{s+1}}.$
        }
    \end{algorithmic}
\end{algorithm}

By Algorithm~\ref{alg:backpropwithcurv}, the computation of the curvature $c(\theta,\dot \theta)$ requires 3 passes in total and the storage of $X$ and $\hat X$ whereas the computation of the gradient requires $2$ passes and the storage of $X$.
Hence the memory footprint is multiplied by $2$ and the computation time by $1.5$. We show in Section~\ref{sec:proof:divide-conquer} how to design a divide-and-conquer algorithm that changes this cost to $(2C,1M)$.

\begin{theorem}\label{thm:computational_time}
    If $(1C,1M)$ represents respectively the computational time and memory footprint of the standard backpropagation method, Algorithm~\ref{alg:backpropwithcurv} costs either $(1.5C,2M)$ or $(2C,1M)$.
\end{theorem}
This result is of importance since it states that computing the exact curvature is at least as cheap as using numerical differences of the gradient \cite{castera2022second}.

\subsection{Proof of Algorithm~\ref{alg:backpropwithcurv}}\label{sec:proof_algocurv}
The goal of this section is to analyse the complexity of computing the curvature term and to prove Algorithm~\ref{alg:backpropwithcurv}.

\paragraph{Forward pass}
We recall that the forward pass is computed through the recurrence
\[
    x_{s+1}(\Theta) =\mathcal F_{s}(x_s(\Theta),\theta_s), \quad 0 \le s\le n.
\]
Moreover the objective function is defined as $\Jc(\Theta)=x_{n+1}(\Theta)$. The computation of $X$ through the recurrence~\eqref{eq:forward:co} is denoted as the {\em forward pass}.

\paragraph{Tangent pass}
Given $\Theta$, a set of data $X(\Theta)$, and an arbitrary direction $\dot \Theta=(\dot \theta_s)_{s=0..n}$, the tangent $\dot X=(\dot x_s)_{s=0..n}$ is defined as 
\[\dot x_s=\lim_{\tau\rightarrow 0} \frac{x_s(\Theta+\tau \dot \Theta)- x_s(\Theta)}{\tau}.\]
For each layer $s$, recall that  $\partial_x \mathcal F_s$ (resp. $\partial_\theta \mathcal F_s$) is the differential of $\mathcal F_s$ with respect to the parameter $x$ (resp. $\theta$) at the point $(x_s(\Theta),\theta_s)$. From now, we omit the notation of the point at which the differential is taken in order to simplify the notations.
By the chain rule theorem, we have that if $\dot x_s$ exists, then the forward recurrence~\eqref{eq:forward:co} yields
\begin{equation}\label{eq:directional:derivative_proof}
    \dot x_{s+1}=(\partial_x \mathcal F_s)\dot x_s+(\partial_\theta \mathcal F_s)\dot \theta_s, \quad \dot x_0=0.
\end{equation}
A recurrence on $s$ allows obtaining existence of $\dot X$ and the scaling
\[X(\Theta +\tau \dot \Theta)=X(\Theta)+\tau \dot X +O(\tau^2).\]
Hence, if $X(\Theta)$ is computed and $\dot \Theta$ is chosen, then $\dot X$ -- the tangent in direction $\dot \Theta$ -- can be computed via the forward recurrence \eqref{eq:directional:derivative_proof} and we have 
\begin{equation}
\label{eq:implicit:gradj}
\langle\nabla \mathcal J (\Theta) ,\dot \Theta\rangle=\dot x_{n+1}.
\end{equation}
The recurrence~\eqref{eq:directional:derivative_proof} which allows the computation of $\dot X$ is coined as the {\em tangent pass}.

\paragraph{Adjoint/backward pass}
In order to compute the gradient, one resorts to the backpropagation algorithm which allows reversing the recurrence~\eqref{eq:directional:derivative_proof} that defines the tangent and computing directly $\hat \Theta=(\hat \theta_s)_{s=0..n}$ such that
\[\langle\nabla \Jc (\Theta) ,\dot \Theta\rangle=\dot x_{n+1} = \sum_s\langle\dot \theta_s,\hat \theta_s\rangle_{\mathcal G_s}.\]
The vector $\hat \Theta$ is then equal to $\nabla \Jc (\Theta)$, provided that one uses the scalar product induced by the sum of the scalar products of all $\mathcal G_s$.
To compute $\hat \Theta$, we use $\partial_x \mathcal F_s^*:\mathcal H_{s+1} \rightarrow \mathcal H_s$ and $\partial_\theta \mathcal F_s^* : \mathcal H_{s+1} \rightarrow \mathcal G_s$ the adjoints of the differentials of $\mathcal F_s$.
The backward of the data $\hat X=(\hat x_s)_{s=1..n+1}$ and the backward-gradient $\hat \Theta=(\hat \Theta_s)_{s=0..n}$ are defined by the reversed recurrence:

\begin{equation}\label{eq:backward:appendix}
    \begin{cases} 
        \hat x_s =(\partial_x \mathcal F_s)^*\hat x_{s+1} \quad \text{ with }\hat x_{n+1}=1 \\
        \hat \theta_s =(\partial_\theta \mathcal F_s)^*\hat x_{s+1}.
    \end{cases}
\end{equation}
The definition of the adjoint and the formula of the tangent~\eqref{eq:directional:derivative_proof} give the following equality:
\begin{align}
    \langle \dot x_{s+1},\hat x_{s+1}\rangle_{\mathcal H_{s+1}}
    &= \langle (\partial_x \mathcal F_s)\dot x_s+(\partial_\theta \mathcal  F_s)\dot \theta_s, \hat x_{s+1}\rangle_{\mathcal H_{s+1}}\nonumber \\
    &= \langle \dot x_s, (\partial_x \mathcal  F_s)^*\hat x_{s+1}\rangle_{\mathcal H_{s+1}}  +\langle \dot \theta_s, (\partial_\theta\mathcal F_s)^* \hat x_{s+1}\rangle_{\mathcal H_{s+1}} \nonumber\\
    &=\langle \dot x_s, \hat x_s\rangle_{\mathcal H_s} +\langle \dot \theta_s, \hat \theta_s\rangle_{\mathcal G_s} \label{eq:inversion:recurrence1}
\end{align}

Summing up the above equations for every $s$, we obtain:
\[\dot x_{n+1}=\langle \dot x_{n+1},\hat x_{n+1}\rangle_{\mathcal H_{n+1}}=\langle \dot x_{0},\hat x_{0}\rangle_{\mathcal H_{n+1}}
+ \sum_s\langle\dot \theta_s,\hat \theta_s\rangle_{\mathcal G_s} = \sum_s\langle\dot \theta_s,\hat \theta_s\rangle_{\mathcal G_s},
\]
where we use $\dot x_0=0$ and $\hat x_{n+1}=1$. We then obtain the celebrated backward propagation formula
\[\nabla \Jc(\Theta) =\hat \Theta.\]

The complexity analysis of the computation of the gradient by the backward formula shows that it requires the computation and the storage of the forward pass in order to be able to evaluate $(\partial_x \mathcal F_s)^*$ and $(\partial_\theta \mathcal F_s)^*$ at the point $(x_s(\Theta),\theta_s)$.

\paragraph{Computing the curvature}
Equation~\ref{eq:implicit:gradj} is the implicit definition of $\nabla \Jc$, where $\dot X$ is defined by the recurrence~\eqref{eq:directional:derivative_proof}. The trick of automatic differentiation is to use the the backward $\hat X$ defined in recurrence~\eqref{eq:backward:appendix} to reverse \eqref{eq:directional:derivative_proof}. This inversion is performed in \eqref{eq:inversion:recurrence1} and it allows not computing the tangent $\dot X$.
We show in this paragraph that the backward $\hat X$ also reverses the recurrence defining the second order term $\ddot X$ defined in~\eqref{eq:defin:secondorder} below.
Once the direction $\dot \Theta$ is chosen, the curvature term can be computed by only a forward pass. To this end, for any direction $\dot \Theta$, introduce $\ddot X=(\ddot x_s)_{s=0..n+1}$ as:
\begin{equation}\label{eq:defin:secondorder}
    \ddot x_s=\lim_{\tau\rightarrow 0} \frac{x_s(\Theta+\tau \dot \Theta)-x_s(\Theta)- \tau\dot x_s}{\tau^2},
\end{equation}
where $\dot x_s$ is the tangent defined in \eqref{eq:directional:derivative_proof}.
Recall that $\nabla^2 \mathcal F_s: \mathcal H_s \times \mathcal G_s \to \mathcal H_{s+1}$ is the bilinear symmetric mapping that represents the second order differentiation of $\mathcal F_s$ at point $(x_s(\Theta),\theta_s)$. It is defined as the only bilinear symmetric mapping that verifies for every $(h_x,h_\theta)$ the relation
\begin{align*}
    \mathcal F_s(x_s(\Theta)+h_x,\theta_s+h_\theta)=&
    \mathcal F_s(x_s(\Theta),\theta_s)
    +\partial_x \mathcal F_s h_x+ \partial_\theta \mathcal F_s h_\theta
    +\frac{1}{2} \nabla^2\mathcal F_s(h_x,h_\theta)\otimes(h_x,h_\theta)\\
    &+o(\Vert h_x\Vert^2+\Vert h_\theta\Vert^2)
\end{align*}
It is easy to prove that $\ddot x_s$ exists and verifies:
\begin{equation}\label{eq:second:order}
    \ddot x_{s+1}=(\partial_x \mathcal F_s)\ddot x_s +\frac 1 2 \nabla^2 \mathcal F_s (\dot x_s,\dot \theta_s)\otimes (\dot x_s,\dot \theta_s), \quad\text{ with } \ddot x_0=0.
\end{equation}
Indeed, denote $\xi_s=x_s(\Theta+\tau \dot \Theta)-x_s(\Theta)- \tau\dot x_s$ so that \[\ddot x_s=\lim_{\tau \rightarrow 0} \frac{\xi_s}{\tau^2},\] we have
\begin{align}
    \xi_{s+1}
    &=\mathcal F_s(x_s(\Theta+\tau \dot \Theta),\theta_s+\tau \dot \theta_s)-\mathcal F_s(x_s(\Theta),\theta_s)-\tau(\partial_x \mathcal F_s)\dot x_s -\tau(\partial_\theta \mathcal F_s)\dot \theta_s \nonumber \\
    &=\mathcal F_s(\xi_s+x_s(\Theta)+\tau \dot x_s,\theta_s+\tau \dot \theta_s)-\mathcal F_s(x_s(\Theta),\theta_s)-\tau(\partial_x \mathcal F_s)\dot x_s -\tau(\partial_\theta \mathcal F_s)\dot \theta_s \nonumber \\
    &= (\partial_x \mathcal F_s)\xi_s +\frac {\tau^2} {2}\nabla^2 \mathcal F_s \left(\frac{\xi_s}{\tau}+\dot x_s,\dot \theta_s\right)\otimes \left(\frac{\xi_s}{\tau}+ \dot x_s,\dot \theta_s\right)+o(\tau^2+\Vert\xi_s\Vert^2).
    \label{eq:xi:noapprox}
\end{align}
By a forward recurrence on~\eqref{eq:xi:noapprox}, starting with $\xi_0=0$, we have that $\xi_s=O(\tau^2)$ so that $\ddot x_s$ exists. Dividing~\eqref{eq:xi:noapprox} by $\tau^2$ and taking the limit yields~\eqref{eq:second:order}.

Upon replacing $\dot X$ by $\ddot X$, the trick used in~\eqref{eq:inversion:recurrence1} can be applied and translates into:
\begin{align*}
    \langle \ddot x_{s+1},\hat x_{s+1}\rangle_{\mathcal H_{s+1}} 
    &= \langle (\partial_x \mathcal F_s)\ddot x_s+\f12 \nabla^2 \mathcal F_s (\dot x_s,\dot \theta_s)\otimes (\dot x_s,\dot \theta_s), \hat x_{s+1}\rangle_{\mathcal H_{s+1}} \\
    &= \langle \ddot x_s, (\partial_x \mathcal  F_s)^*\hat x_{s+1}\rangle_{\mathcal H_{s+1}} + \f12\langle \nabla^2 \mathcal F_s (\dot x_s,\dot \theta_s)\otimes (\dot x_s,\dot \theta_s), \hat x_{s+1}\rangle_{\mathcal H_{s+1}}  \\
    &=\langle \ddot x_s, \hat x_s\rangle_{\mathcal H_s} +\f12  \langle \nabla^2 \mathcal F_s (\dot x_s,\dot \theta_s)\otimes (\dot x_s,\dot \theta_s), \hat x_{s+1}\rangle_{\mathcal H_{s+1}}
\end{align*}
Summing up these equations in $s$ and using $\ddot x_0=0$ and $\hat x_{n+1}=1$, we obtain
\begin{align*}
     \ddot x_{n+1}&=\langle \ddot x_{n+1},\hat x_{n+1}\rangle_{\mathcal H_{n+1}} \\
    &=\langle \ddot x_{0},\hat x_{0}\rangle_{\mathcal H_{n+1}}
    + \sum_s\f12 \langle \nabla^2 \mathcal F_s (\dot x_s,\dot \theta_s)\otimes (\dot x_s,\dot \theta_s), \hat x_{s+1}\rangle_{\mathcal H_{s+1}} \\
    & = \sum_s\f12 \langle \nabla^2 \mathcal F_s (\dot x_s,\dot \theta_s)\otimes (\dot x_s,\dot \theta_s), \hat x_{s+1}\rangle_{\mathcal H_{s+1}}.
\end{align*}
In order to conclude and prove Algorithm~\ref{alg:backpropwithcurv}, it is sufficient to remark that $\mathcal J(\Theta)=x_{n+1}(\Theta)$ so that
$\f12 \langle\nabla^2 \mathcal J(\Theta)\dot \Theta,\dot \Theta\rangle=\ddot x_{n+1}$.

\paragraph{More on automatic differentiation}
In Section~\ref{sec:Pearlmutter}, the reader will find a method to compute the matrix-vector product with the Hessian. This method is not new and is known as the {\em Pearlmutter's trick}  \cite{christianson1992automatic,pearlmutter1994fast}. We prove this trick in our setting in order to link our computations with other automatic-differentiation techniques.
Moreover, we also give some of the expression of $\nabla^2 \mathcal F_s (\dot x_s,\dot \theta_s)\otimes (\dot x_s,\dot \theta_s)$ for standard layers in Appendix~\ref{sec:structlayers} to settle the notations.

\subsection{Proof of Theorem~\ref{thm:computational_time}}\label{sec:proof:divide-conquer}

Recall that $(1C,1M)$ is the complexity of a gradient computation, we show how to change the overall cost of computing the curvature from $(1.5C,2M)$ to $(2C,1M)$ by a divide-and-conquer algorithm.
In order to simplify the analysis, several simplifications are made.
\begin{itemize}
    \item There are three kind of passes, the forward pass in \eqref{eq:forward:co} that computes $X$, the backward pass in \eqref{eq:backward:appendix} that computes $\hat X$ and $\hat \Theta$ and the tangent-curvature pass described in Algorithm~\ref{alg:backpropwithcurv} that computes the curvature. We suppose that each of these passes have roughly the same computational cost $C/2$. This assumption is subject to discussion. In one hand the backward and tangent passes require each twice as much matrix multiplication as the forward pass. On the other hand, soft activation functions are harder to compute in the forward pass.
    \item We assume that storing $X$ or $\hat X$ has the same memory footprint $1M$. Notably, we suppose that the cost of storing the parameters $\Theta$ or the gradient $\hat \Theta$ is negligeable with respect to the storage of the data through the network. This assumption can only be made for optimization with large enough batches $\Bc_k$.
    \item We suppose that we can divide the neural network in two pieces that each costs half the memory and half the computational time. This means that we are able to find $L$, such that the storage of $(x_s)_{s \le L}$ and the storage of $(x_s)_{s \ge L}$ have same memory footprint $M/2$. Moreover we suppose that performing a pass for $s\ge L$ or for $s\le L$ costs $C/4$ computational time. This assumption is reasonable and simplifies the analysis but it is of course possible to exhibit pathological networks that won't comply with this assumption.
    \item We suppose that the only cost in data transfer comes from the initialization of the parameters $\Theta$, the initial data $x_0$ and the direction of descent $\dot \Theta$. Note that the computation of $\dot \Theta$ requires the computation of the gradient $\hat \Theta$.
\end{itemize}

\pgfmathsetmacro{\rectwidth}{0.6}
\pgfmathsetmacro{\rectheight}{3}

\pgfmathsetmacro{\hc}{0.15}
\pgfmathsetmacro{\hx}{\rectwidth+\hc+0.5}
\pgfmathsetmacro{\hy}{0}
\pgfmathsetmacro{\hdeco}{0.4}

\definecolor{colorforw}{rgb}{0.6, 0., 0}
\definecolor{colorback}{rgb}{0, 0.6, 0.}
\definecolor{colorcurv}{rgb}{0., 0, 0.6}
\definecolor{colortitle}{rgb}{0.4, 0.4, 0.4}

\tikzstyle{nodestyle}=[draw,circle,minimum width=0.8*\rectwidth,minimum height=\rectwidth,very thick,font=\bfseries]
\tikzstyle{arrowstyle}=[-latex,very thick]
\tikzstyle{rect}=[rounded corners=2]

\newcommand{\deco}[1]{
\pgfmathsetmacro{\decothick}{0.07}

\draw[color=white,fill=colortitle,rounded corners=6] (-\decothick-\hdeco,-\decothick-1.3*\rectwidth-\hdeco) rectangle (\decothick+2*\hx+\rectwidth+\hdeco,\decothick+2*\hy+\rectheight+\hdeco);
\draw[color=white,fill=white,rounded corners=6] (-\hdeco,-1.3*\rectwidth-\hdeco) rectangle (2*\hx+\rectwidth+\hdeco,2*\hy+\rectheight+\hdeco);
\node[color=colortitle,fill=white] (t) at (0.7,-1.3*\rectwidth-\hdeco) {#1};
}

\newcommand{\forw}[4]{
    \node[nodestyle,color=colorforw,fill=colorforw!10] (fw) at (0.5*\rectwidth,-0.7*\rectwidth)  {F};
    \draw[rect] (0,0) rectangle (\rectwidth,\rectheight);
    \ifthenelse{-1< #1}{\draw[fill=colorforw,rect] (0,\rectheight-#1*\rectheight) rectangle (\rectwidth,\rectheight-#2*\rectheight);}{}
    
    \ifthenelse{-1< #3}{\draw [color=colorforw,arrowstyle]  (\rectwidth+\hc,\rectheight-#3*\rectheight)-- (\rectwidth+\hc,\rectheight-#4*\rectheight);}{}
}

\newcommand{\back}[4]{
    \node[nodestyle,color=colorback,fill=colorback!10] (fw) at (\hx+0.5*\rectwidth,\hy+-0.7*\rectwidth)  {B};
    \draw[rect] (\hx,\hy) rectangle (\hx+\rectwidth,\hy+\rectheight);
    \ifthenelse{-1< #1}{\draw[fill=colorback,rect] (\hx,\hy+\rectheight-#1*\rectheight) rectangle (\hx+\rectwidth,\hy+\rectheight-#2*\rectheight);}{}
    \ifthenelse{-1< #3}{\draw [color=colorback,arrowstyle]  (\hx+\rectwidth+\hc,\hy+\rectheight-#4*\rectheight)-- (\hx+\rectwidth+\hc,\hy+\rectheight-#3*\rectheight);}{}
}

\newcommand{\curv}[4]{
    \node[nodestyle,color=colorcurv,fill=colorcurv!10] (fw) at (2*\hx+0.5*\rectwidth,2*\hy+-0.7*\rectwidth)  {T};
    \draw[rect] (2*\hx,2*\hy) rectangle (2*\hx+\rectwidth,2*\hy+\rectheight);
    \ifthenelse{-1< #1}{\draw[fill=colorcurv,rect] (2*\hx,2*\hy+\rectheight-#1*\rectheight) rectangle (2*\hx+\rectwidth,2*\hy+\rectheight-#2*\rectheight);}{}
    \ifthenelse{-1< #3}{\draw [color=colorcurv,arrowstyle]  (2*\hx+\rectwidth+\hc,2*\hy+\rectheight-#3*\rectheight)-- (2*\hx+\rectwidth+\hc,2*\hy+\rectheight-#4*\rectheight);}{}
}
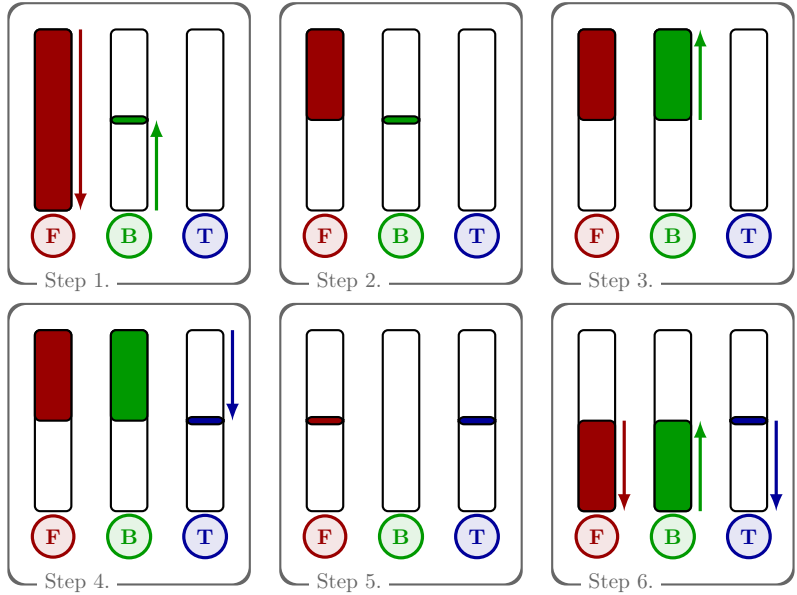
\begin{figure}
    \centering
    \begin{tikzpicture}[thick,scale=0.8, every node/.style={transform shape}]
        \deco{Step 1.}
        \forw{0.}{1.}{0.}{1.}
        \back{0.48}{0.52}{0.5}{1.}
        \curv{-2}{0.}{-2}{0.}
    \end{tikzpicture}\hspace{0.2cm}
    \begin{tikzpicture}[thick,scale=0.8, every node/.style={transform shape}]
        \deco{Step 2.}
        \forw{0.}{0.5}{-2.}{1.}
        \back{0.48}{0.52}{-2}{1.}
        \curv{-2}{0.}{-2}{0.}
    \end{tikzpicture}\hspace{0.2cm}
    \begin{tikzpicture}[thick,scale=0.8, every node/.style={transform shape}]
        \deco{Step 3.}
        \forw{0.}{0.5}{-2.}{1.}
        \back{0.}{0.5}{0}{0.5}
        \curv{-2}{0.}{-2}{0.}
    \end{tikzpicture}\\
    \begin{tikzpicture}[thick,scale=0.8, every node/.style={transform shape}]
        \deco{Step 4.}
        \forw{0.}{0.5}{-2.}{1.}
        \back{0.}{0.5}{-2}{0.5}
        \curv{0.48}{0.52}{0.}{0.5}
    \end{tikzpicture}\hspace{0.2cm}
    \begin{tikzpicture}[thick,scale=0.8, every node/.style={transform shape}]
        \deco{Step 5.}
        \forw{0.48}{0.52}{-2.}{1.}
        \back{-2.}{0.5}{-2}{0.5}
        \curv{0.48}{0.52}{-2}{0.5}
    \end{tikzpicture}\hspace{0.2cm}
    \begin{tikzpicture}[thick,scale=0.8, every node/.style={transform shape}]
        \deco{Step 6.}
        \forw{0.5}{1.}{0.5}{1.}
        \back{0.5}{1}{0.5}{1.}
        \curv{0.48}{0.52}{0.5}{1.}
    \end{tikzpicture}
    \caption{Illustration of the divide-and-conquer algorithm that changes the cost of computating the curvature from $(1.5C,2M)$ to $(2C,1M)$.
    The rectangles above the letters {\bf F, B, T} represent the three different passes (in order: forward, backward and tangent).
    The memory usage is represented by color-filling in the rectangles, the computations are represented by arrows on the right of the passes.
    In total, the filled area never exceeds $1$ rectangle, hence memory usage is $1M$.
    The total length of the arrows is $4$ times the length of a rectangle, this represents $4$ passes.
    The computational time is then twice the computational time of the standard backward algorithm.
    }
    \label{fig:illustration_divideconquer}
\end{figure}

We now describe how to compute the curvature with $(2C,1M)$ and no extra data transfer. We display the current memory load and the elapsed computational time at the end of each phase.
A visual illustration of this algorithm is proposed in Figure~\ref{fig:illustration_divideconquer}.
\begin{enumerate}\setcounter{enumi}{-1}
    \item Transfer the data $x_0$ and $\Theta$. 
    \item Compute $X=(x_s)_{s}$ and store it.  For $s\ge L$, compute the  backward via~\eqref{eq:backward} without storing it. \mbox{} \hfill Cost is $(\frac 3 4 C,1M)$
    \item Flush from memory $(x_s)_{s\ge L}$. \hfill Cost is $(\frac 3 4 C,\f12 M)$ 
    \item For $s \le L$, compute the backward via~\eqref{eq:backward} and store it. \hfill Cost is $(1C,1M)$
    \item Choose the descent direction and transfer the data $\dot \Theta$. Compute the tangent via~\eqref{eq:directional:derivative_proof} for $s\ge L$. \mbox{} \hfill Cost is $(\frac 5 4 C,M)$
    \item  Flush from memory $(\hat x_s)_{s}$ and $(x_s)_{s<L}$. \hfill Cost is $(\frac 5 4 C,0M)$
    \item For $s\ge L$, compute the forward, the backward and store them. Compute the tangent for $s\ge L$. \mbox{} \hfill Cost is $(2 C,1M)$
\end{enumerate}

\section{Numerical experiments}\label{sec:num}

\subsection{Convergence/exploration trade off}\label{sec:conv_expl_tradeoff}

\subsubsection{The RED algorithm}
In order to test the convergence/exploration trade-off, we reproduce the benchmark of \cite{castera2022second}. We set ourselves in the case where the initial parameters are randomly chosen, so that the practitioner wants a smooth transition from exploration ($\ell=1$) to convergence ($\ell=\frac 1 2$). We choose in Algorithm~\ref{alg:RED} a simple, per epoch, exponential decay rule of the learning rate $\ell$ from $1$ to $1/2$. This algorithm is coined as RED (Rescaled with Exponential Decay). We purposely unplug any other tricks of the trade, notably Robbins-Monro convergence conditions. Indeed, a Robbins-Monro decay rule would interfere with our analysis. Algorithm RED is not a production algorithm, it serves at testing the ``natural'' convergence properties of rescaling.
In Appendix~\ref{sec:comparisonexistingalgo_appendix}, we provide a comparison of RED with a standard SGD that has Robbins-Monro decaying conditions.
Due to the remark in Section~\ref{sec:analysis_rescaling}, we make clear that $L^2$-regularization is used. If $\Theta\mapsto \mathcal L_s(\Theta)$ is the original loss function, then the function $\mathcal J_s$ is defined as $\mathcal J_s(\Theta) \eqdef \mathcal L_s(\Theta) +\frac \lambda 2 \Vert \Theta\Vert^2$.

\begin{algorithm}
\caption{RED (rescaled-exponential-decay) for SGD or RMSProp preconditioning {\bf and no convergence guaranty}}
\begin{algorithmic}[1]
\State {\bf Input parameters} $\beta_2=0.999$ (RMSProp parameter), RMSProp (boolean), $\lambda > 0$ ($L^2$-regularization), $N$ (total number of epochs), $\varepsilon=10^{-8}$ (numerical stabilization).
\State {\bf Initialization} $\hat v_0=0$, $\Theta_0$ random, $\ell=1$ initial learning rate and $\eta=1/2$ the step multiplicative factor between the first and the last iterations. 
\For{$k=1,2,..$}
  \State $g_k=\E_{s \in \Bc_k}\left[\nabla \Jc_s(\Theta_k)\right]$ \Comment{gradient} \label{lst:line:start-ema-g}
  \If{RMSProp} \Comment{RMSProp preconditioning} \label{lst:line:start-precond}
  	\State $\hat v_k=\beta_2\hat v_{k-1}+(1-\beta_2) g^2_k\quad$ and $\quad\tilde v_k=\hat v_{k}/(1-\beta_2^k)$ and $P_k=\diag(\sqrt{\tilde v_k}+\varepsilon)$
  \Else
    \State $P_k=\Id$
  \EndIf \label{lst:line:end-precond}
  \State $\dot \Theta_k=P_k^{-1}g_k$ \Comment{direction of update} \label{lst:line:dir_descent}
  \State {\bf Use Algorithm~\ref{alg:rescale} and compute $r_k$} \Comment{rescaling} \label{lst:line:rescale}
  \State $\Theta_{k+1}=\Theta_k - \ell r_k \dot \Theta_k$  \Comment{parameters update}   \label{lst:line:step}
  \State At the end of each epoch $\ell \leftarrow  \eta^{\frac 1  {N}}\ell$ \label{lst:line:decrease_ell}
  \EndFor
\end{algorithmic}
\label{alg:RED}
\end{algorithm}

The numerical experiments are done on the benchmark of ~\cite{castera2022second}. It consists in four test cases, a MNIST classifier \cite{lecun2010mnist}, a CIFAR-10 classifier \cite{krizhevsky2009learning} with VGG11 \cite{simonyan2014very} architecture, a CIFAR-100 classifier with VGG19 and the classical autoencoder of MNIST described in \cite{hinton2006reducing}.
The ReLU units are replaced by smooth versions in order to compute the curvature term, and $L^2$ regularization is added to each test.
The models are trained with a batch size of $256$ and the number of epochs is set to $200$ for MNIST classification and $500$ for the others.
The precise set of parameters that allows reproductibility is described in Appendix~\ref{sec:xpdescription}.
We also give indications of the computational time on an NVIDIA Quadro RTX 5000.
Each experiment is run $3$ times with different random seeds and we display the average of the tests with a bold line, the limits of the shadow area are given by the maximum and the minimum over the runs.
When displaying the training loss or the step histories, an exponential moving average with a factor $0.99$ is applied in order to smooth the curves and gain in visibility.
Note that the training and testing loss functions are displayed with the $L^2$ regularization term.
On all figures the $x$-axis is the number of epochs. Remember however that the computational cost is not the same for the different optimizers, see Theorem~\ref{thm:computational_time}.

\subsubsection{Interpretation of the RED experiments}\label{sec:autovsmanual}

\begin{figure}
    \centering
    \def\figwidth{.24\linewidth}
    \begin{tabular}{@{}c@{}c@{}c@{}c@{}c@{}}
        \rotatebox[origin=c]{90}{\small Training loss} &
        \includegraphics[valign=m,width=\figwidth]{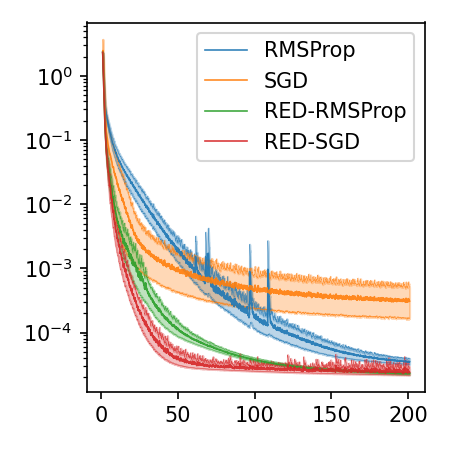} &
        \includegraphics[valign=m,width=\figwidth]{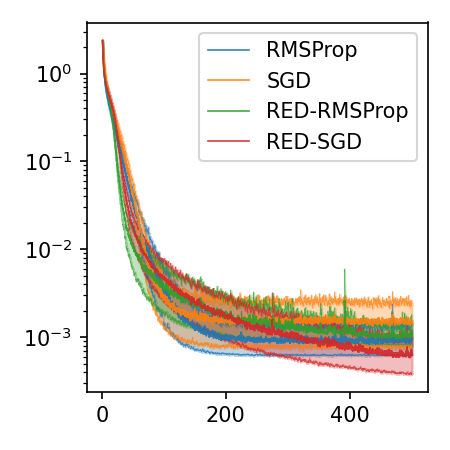} &
        \includegraphics[valign=m,width=\figwidth]{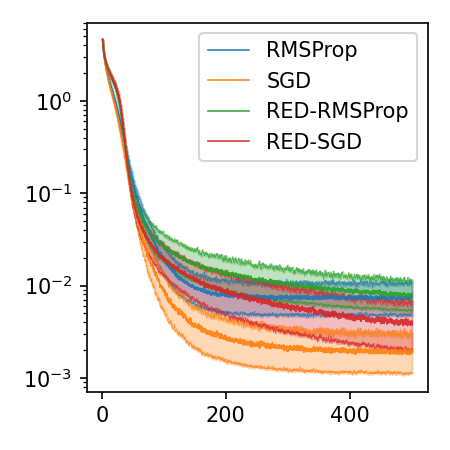} &
        \includegraphics[valign=m,width=\figwidth]{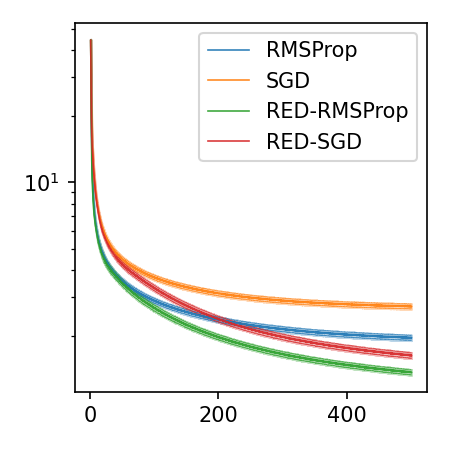}
        \\

        \rotatebox[origin=c]{90}{\small Step} &
        \includegraphics[valign=m,width=\figwidth]{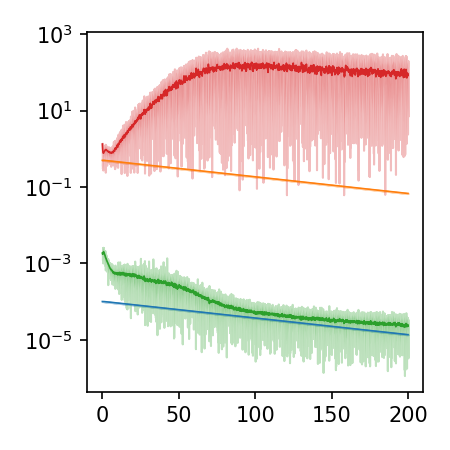} &
        \includegraphics[valign=m,width=\figwidth]{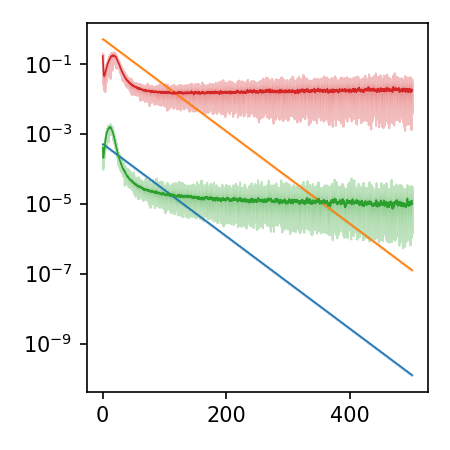} &
        \includegraphics[valign=m,width=\figwidth]{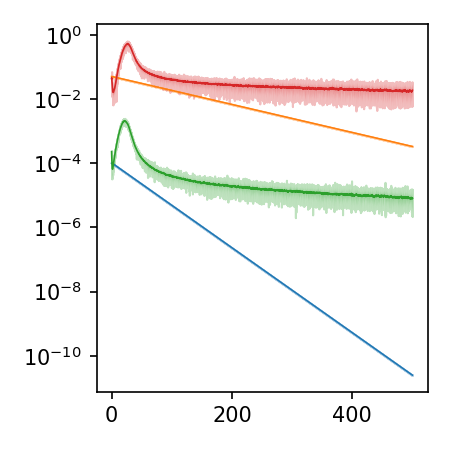} &
        \includegraphics[valign=m,width=\figwidth]{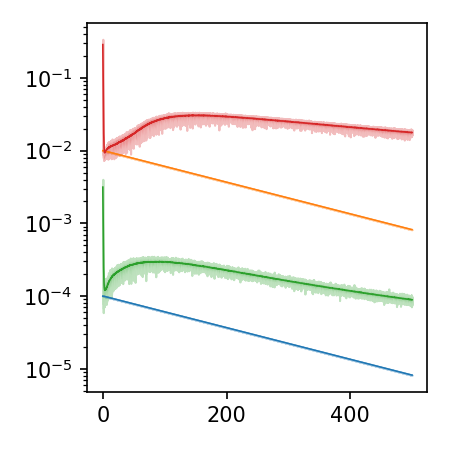}
        \\

        \rotatebox[origin=c]{90}{\small Testing loss} &
        \includegraphics[valign=m,width=\figwidth]{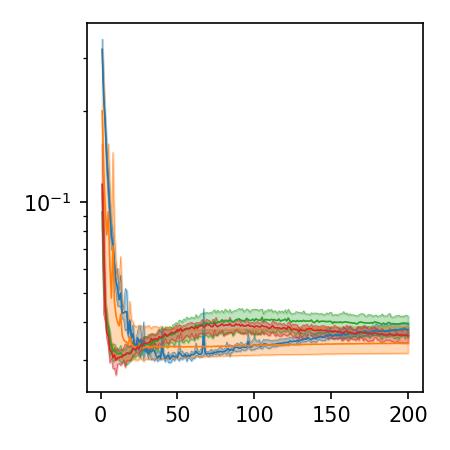} &
        \includegraphics[valign=m,width=\figwidth]{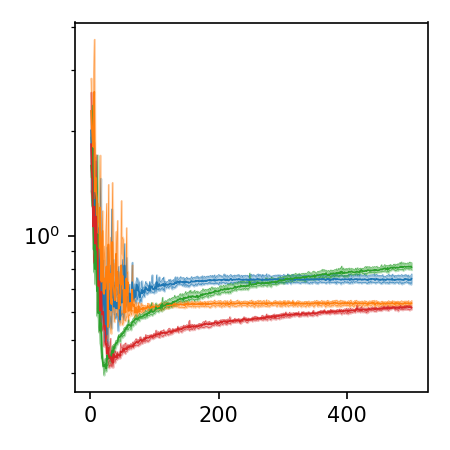} &
        \includegraphics[valign=m,width=\figwidth]{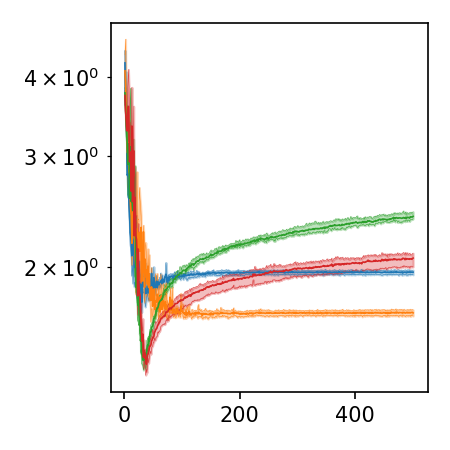} &
        \includegraphics[valign=m,width=\figwidth]{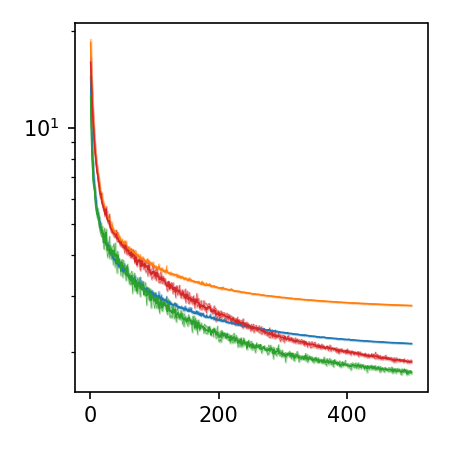}
        \\

        \rotatebox[origin=c]{90}{\small Accuracy test} &
        \includegraphics[valign=m,width=\figwidth]{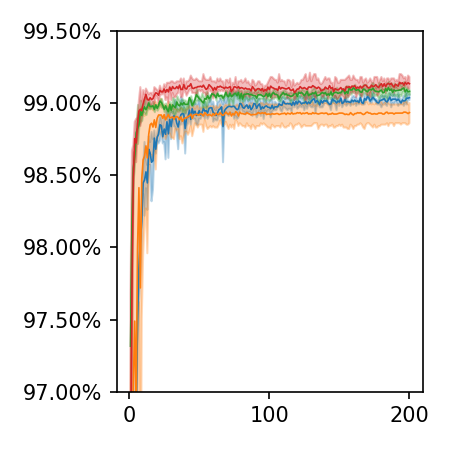} &
        \includegraphics[valign=m,width=\figwidth]{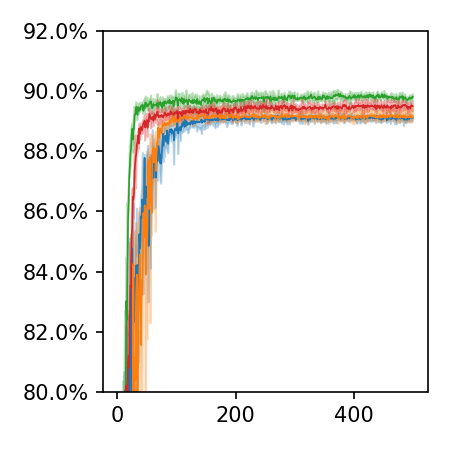} &
        \includegraphics[valign=m,width=\figwidth]{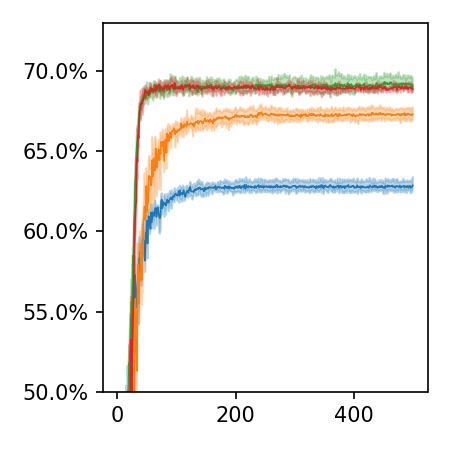} &
        \hspace{.24\linewidth}
        \\

        & MNIST &
        CIFAR10 &
        CIFAR100 &
        autoencoder
    \end{tabular}
    \figspace
    \caption{Training loss, step size, testing loss and test accuracy for the RED and manually-tuned SGD and RMSProp optimizers.
    Each column gives the different test cases (resp. MNIST, CIFAR10, CIFAR100 and autoencoder).
    The RED method which has no tuning gives competitive results in comparison with the manually-tuned SGD and RMSProp optimizers.
    }
    \label{fig:autovsmanual}
\end{figure}

In this first set of experiments, we compare the RED method given in Algorithm~\ref{alg:RED} with standard SGD and RMSProp.
In order to recover these two latter algorithms, set $r_k=1$ in line~\ref{lst:line:rescale} of Algorithm~\ref{alg:RED}. The hyperparameters, namely the initial learning rate $\ell$ and its decay factor $\eta$, are optimized on the training loss with a grid search over the $20\%$ first epochs, these algorithms are coined as ``standard algorithms''.
The results are displayed in Figure~\ref{fig:autovsmanual} for the standard algorithms (orange for SGD, blue for RMSProp) and their RED version (red for SGD, green for RMSProp).

\paraspace
\paragraph{Training loss} The analysis of the training loss shows that RED is competitive to the standard SGD and RMSProp methods.
Note however that the hyperparameters of the standard methods have been chosen as to optimize the behavior of the training loss, hence we cannot expect the RED method to outperform the manually-tuned methods.

\paraspace
\paragraph{Step}
We always observe an increase in the step for the first few epochs ($50$ for MNIST, $10$ for CIFAR). This step increase coincides with the important decrease of the training and testing loss functions.
We interpret this behavior as a search for a basin of attraction of a local minimum. 
It should be noted that the step of the standard CIFAR100 and autoencoder is an order of magnitude smaller than their RED counterpart. Indeed larger steps on these methods cause the algorithm to diverge.
This seems to indicate that the stage of the first $10$ epochs where the step is small is of importance and is well captured by the RED algorithm.  Note that this behavior is the one that is implemented when using warm-up techniques~\cite{loshchilov2016sgdr}. The analysis of the step seems to showcase the power of adaptive rescaling and indicate that warm-up techniques can be handled by the rescaling. This potential is investigated in Section~\ref{sec:hyperexpl}.

\paraspace
\paragraph{Testing loss and accuracy}
The rescaling aims at minimizing quickly the training loss, no conclusions can be drawn from the analysis of the test dataset. Nevertheless, on the CIFAR experiments, an overfitting phenomenon starts from the $25$\textsuperscript{th} epoch approximatively. The overfitting is clearer and more pronounced on the RED method.
This is in accordance with the analysis of the step size: the rescaled method seems to have converged to the maximum of the expressivity of the network at the $50$\textsuperscript{th} epoch.
Concerning the accuracy, it is well known that adaptive methods have poor generalization performances in the overparameterized setting in comparison to SGD \cite{wilson2017marginal}.
Indeed the standard RMSProp achieves lower performance on the test dataset of CIFAR100. Surprisingly, the RED-RMSProp algorithm does not have this property.

As a conclusion of these tests, RED, which is a naive implementation of convergence/exploration trade-off works surprisingly well on this benchmark. We purposely disconnected Robbins-Monro decay rule and let the algorithm run way past overfitting. It still exhibits good convergence properties.

\subsubsection{Other numerical tests}\label{sec:morenumtests}

 In Appendix~\ref{sec:momentum} we investigate the use of momentum with SGD and Adam on the CIFAR100 classifier. The proposed method is only available to deal with direction of descent and the directions of update given by momentum based algorithms are not necessarily direction of descent, yielding poor convergence results.
 
In Appendix~\ref{BS-CIFAR}, we perform tests with smaller batches and we exhibit pathological cases where the rescaled method is highly impacted by stochasticity. The main conclusion is that the performance of the method collapses when the batch is too small compared to the number of classes. This problem in the curvature computation arises at the last layer of the neural network (linear classifier).

In Appendix~\ref{sec:regularization_effect} we study the effet of the $L_2$ regularization on the CIFAR10 classifier, showing numerically that the potential theorical issues raised in Section~\ref{sec:analysis_rescaling} do not impede convergence.  

Finally, in Appendix~\ref{sec:comparisonexistingalgo_appendix} we perform some comparisons with an existing BB method \cite{castera2022second} and with a SGD with Robbins-Monro decay condition.

\subsection{Hyperexploration mode}\label{sec:hyperexpl}
\begin{figure}
    \centering
    \def\figwidth{.24\linewidth}
    \begin{tabular}{@{}c@{}c@{}c@{}c@{}c@{}}
        \rotatebox[origin=c]{90}{\small CIFAR-10} &
        \includegraphics[valign=m,width=\figwidth]{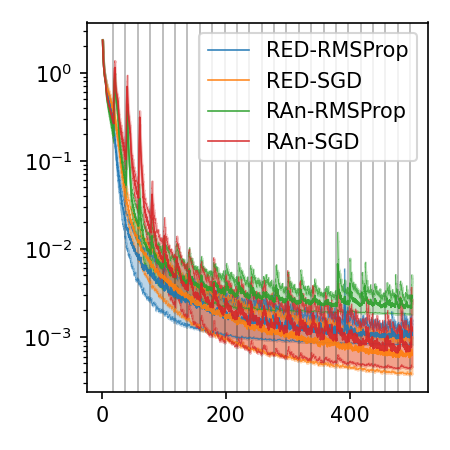}&
        \includegraphics[valign=m,width=\figwidth]{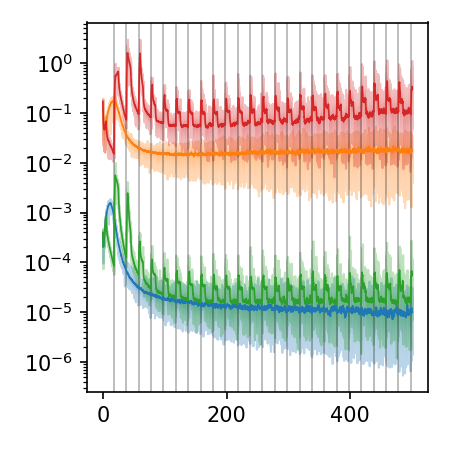}&
        \includegraphics[valign=m,width=\figwidth]{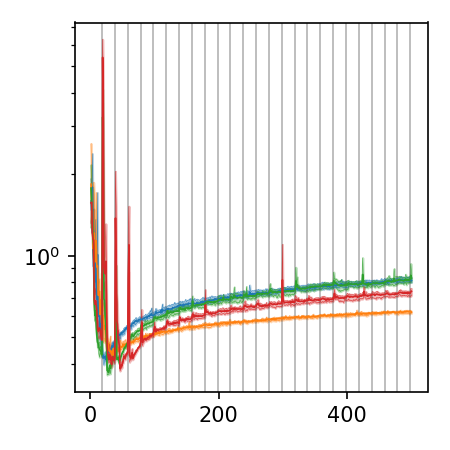}&
        \includegraphics[valign=m,width=\figwidth]{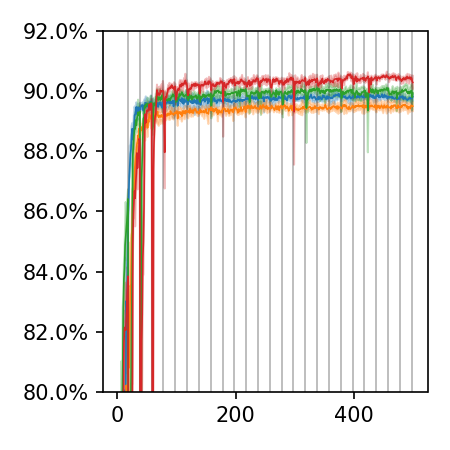}
        \\
        \rotatebox[origin=c]{90}{\small CIFAR-100} &
        \includegraphics[valign=m,width=\figwidth]{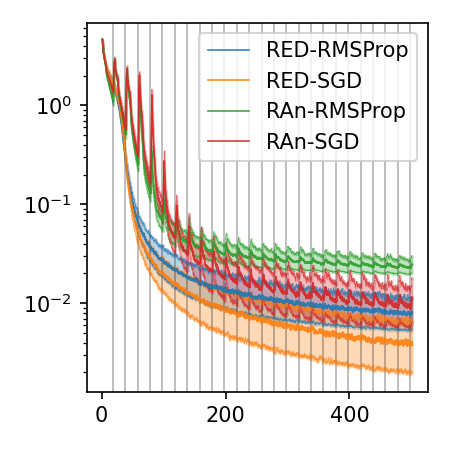}&
        \includegraphics[valign=m,width=\figwidth]{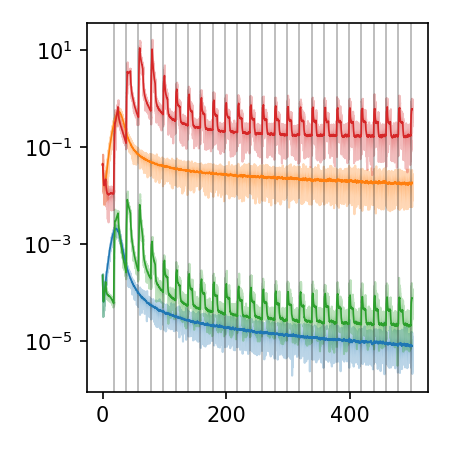}&
        \includegraphics[valign=m,width=\figwidth]{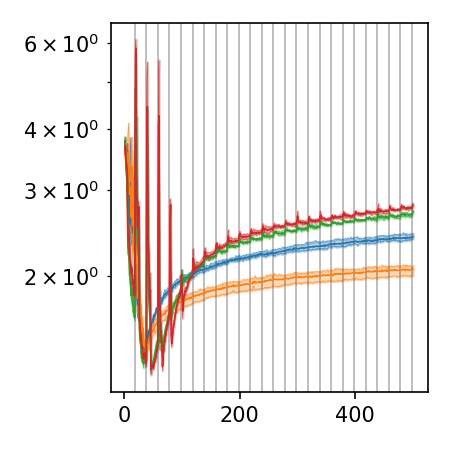}&
        \includegraphics[valign=m,width=\figwidth]{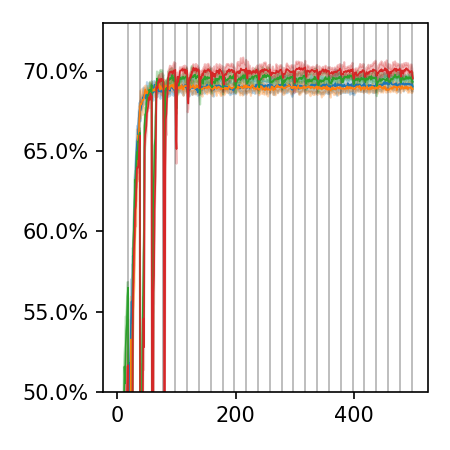}
        \\
         & Training loss &
        Step&
        Testing loss&
        Accuracy test
	\end{tabular}
    \caption{Annealing (RAn) vs Exponential decay (RED) method. The annealing method increases the loss functions during the hyperexploration ($\ell=2$) phase (after the vertical gray lines). This empirically proves that the factor $\ell=1$ is the limiting factor that allows exploration without increasing the loss function. The basin of attraction of the RAn method is different of the one of RED, except possibly for CIFAR10 with RMSProp.}
    \label{fig:annealing}
\end{figure}

In order to showcase hyperexploration, we propose a vanilla annealing method. We replace in Algorithm~\ref{alg:RED} (RED) the line~\ref{lst:line:decrease_ell} (update of the parameter $\ell$) by setting periodically $\ell=1$ for 5 epochs, $\ell=\frac 1 2$ for 13 epochs and $\ell=2$ for 2 epochs. These three phases are coined respectively as \emph{exploration}, \emph{convergence} and \emph{hyperexploration}. We favor sharp changes when letting $\ell$ oscillate in order to easily interpret the results. This simple annealing method is coined RAn (Rescaled Annealing). We display in Figure~\ref{fig:annealing} the results for CIFAR10 and CIFAR100. On Figure~\ref{fig:annealing} the shift between the choice $\ell=\frac 1 2$ and $\ell=2$ is represented by a vertical gray line. We also display the results for the RED algorithm for comparaison. Of importance in Figure~\ref{fig:annealing} is the behavior of the loss function. The latter increases at each \emph{hyperexploration} phase, and converges during the \emph{exploration} and \emph{convergence} phase. A similar effect is also present but less pronounced on the testing loss and accuracy. The increase of the training loss function for $\ell =2$ is in accordance with the theory, and is at the core of annealing methods that aim at escaping local minima. These tests validate the fact that $\ell=1$ is an upper-bound for the \emph{exploration} choice.

\subsection{Hyperconvergence mode}\label{sec:hyperconv}

In this example, we wish to study a more realistic dataset for which stochastic issues are of essence. To that end, we use the ImageNet 1K database and load a state-of-the-art pretrained ResNet-50. This network achieves a $80.858 \%$ top-1 accuracy and a $95.434\%$ top-5 accuracy.
From the study of Appendix~\ref{BS-CIFAR}, summarized in Section~\ref{sec:morenumtests}, we know that important stochastic problems will occur in the last layer (Linear Classifier or LC)  of the DNN. Hence, we erase the parameters of the linear classifier and aim at re-training it while freezing the weights of the feature extractor (upstream section of the network). This setting is reminiscent of a toy {\em transfer learning} problem and aims at training a simple neural network with a state-of-the-art dataset.

From the coupon collector's problem with $1000$ classes, we know that the expectation of $T$, the smallest batch size that obtains at least one element in each class, is approximatively 7.3K when classes are drawn independently and uniformly. We expect stochastic issues to appear when batches are of size smaller than 7.3K. The batch size used to pretrain the network is 1K, hence we test the rescaling for batch of size 1K, 2K, 4K, 8K and 16K. Since stochastic effects should be seriously mitigated for the 8K and 16K cases, these two cases represent a baseline for the training.

\begin{figure}
    \centering
    \def\figwidth{.24\linewidth}
    \begin{tabular}{@{}c@{}c@{}c@{}c@{}c@{}}
        \rotatebox[origin=c]{90}{\small plain training} &
        \includegraphics[valign=m,width=\figwidth]{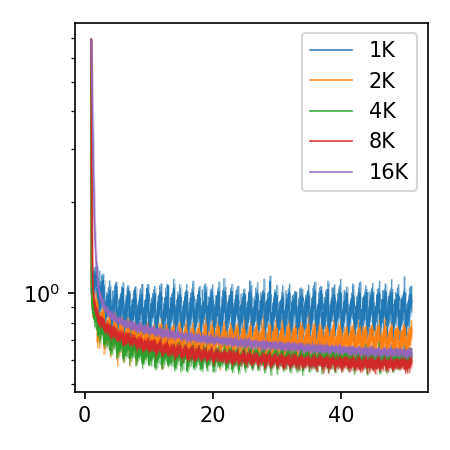} &
        \includegraphics[valign=m,width=\figwidth]{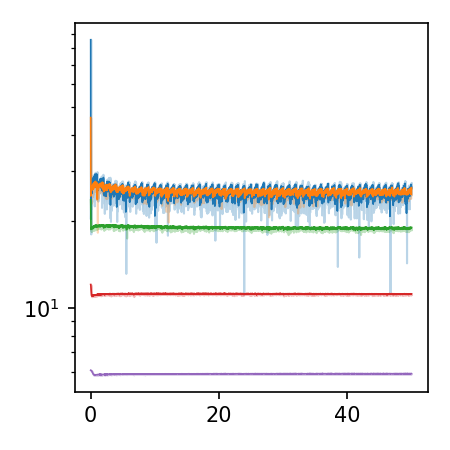} &
        \includegraphics[valign=m,width=\figwidth]{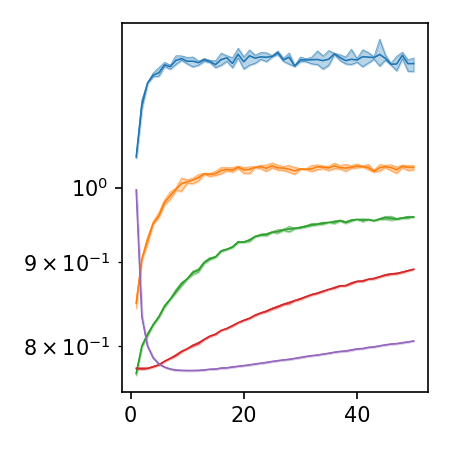} &
        \includegraphics[valign=m,width=\figwidth]{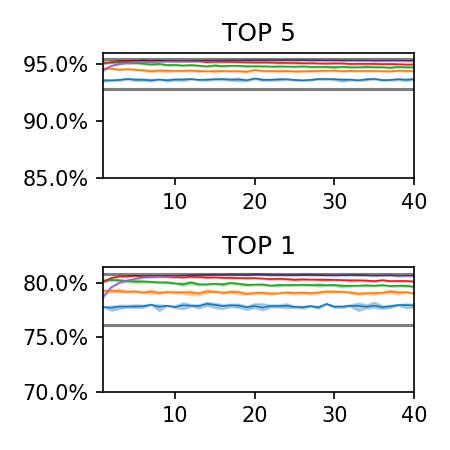} \\
        \rotatebox[origin=c]{90}{\small RA+LS} &
        \includegraphics[valign=m,width=\figwidth]{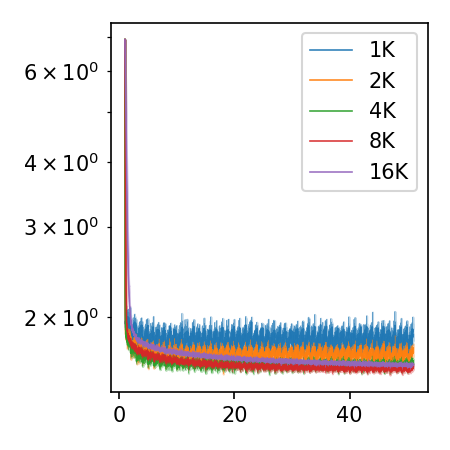} &
        \includegraphics[valign=m,width=\figwidth]{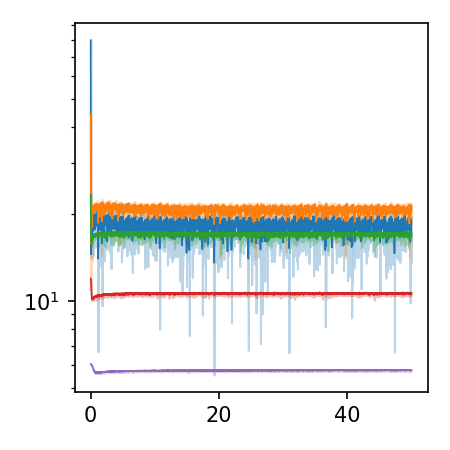} &
        \includegraphics[valign=m,width=\figwidth]{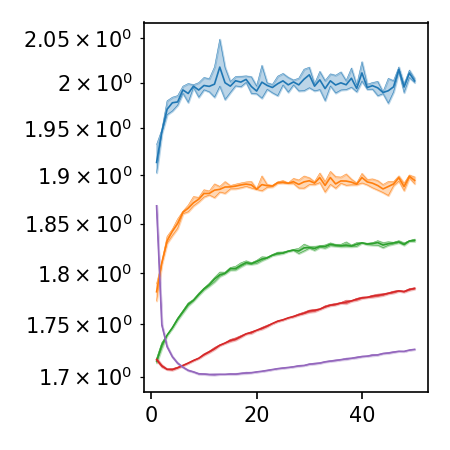} &
        \includegraphics[valign=m,width=\figwidth]{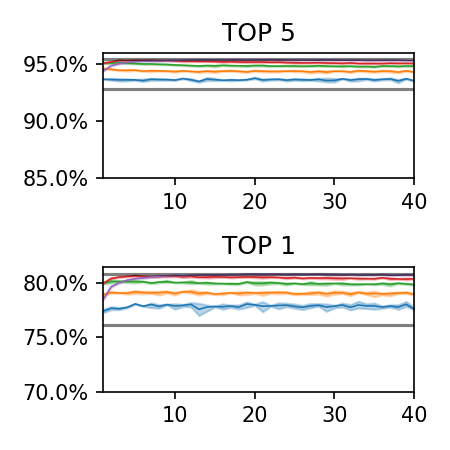} \\
        \rotatebox[origin=c]{90}{\small Cos+RA+LS} &
        \includegraphics[valign=m,width=\figwidth]{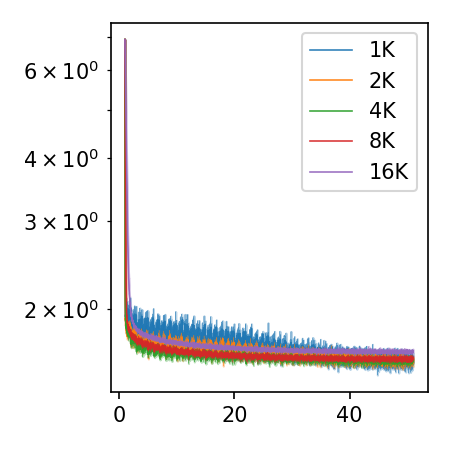} &
        \includegraphics[valign=m,width=\figwidth]{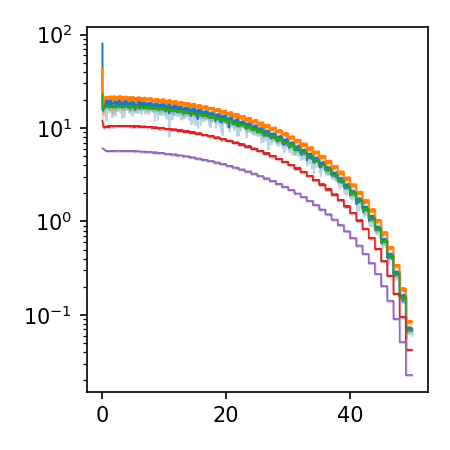} &
        \includegraphics[valign=m,width=\figwidth]{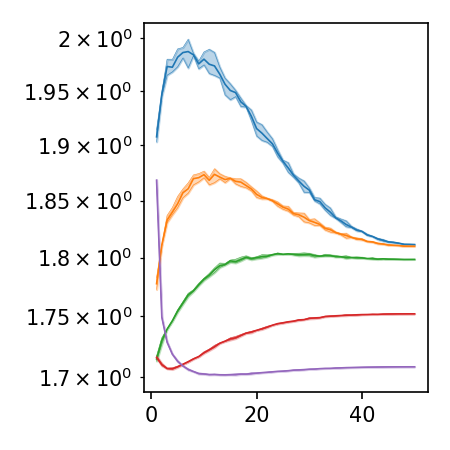} &
        \includegraphics[valign=m,width=\figwidth]{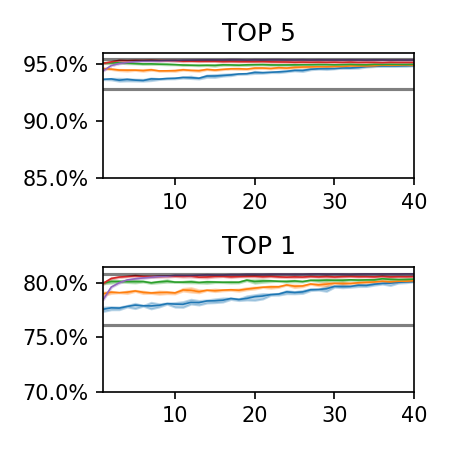} \\

        & Training loss &
        Step &
        Testing loss &
        Accuracy
    \end{tabular}
    \figspace
    \caption{Training a linear classifier on ImageNet 1K with a ResNet-50 feature extractor and with different batch size for SGD.}
    \label{fig:I1K}
\end{figure}

We first discard every trick and test the rescaling for the different batch size. We adopt a fixed rescaled learning rate strategy of $\ell=\f12$ in order to converge as fast as possible. The result is given in Figure~\ref{fig:I1K} top line and referred as {\em plain training}. Of importance in the top line of Figure~\ref{fig:I1K} are the oscillations in the training loss, which are less pronounced as the batch size increases. Note also the stability of the top-1 and top-5 accuracies around a value that depends on the batch size. For the 16K experiment, the linear classifier achieves the top-1 and top-5 accuracies of the pre-trained weigths.

We then implement several tricks of the trade, namely repeated augmentation (RA)~\cite{hoffer2019augment} and label smoothing (LS)~\cite{szegedy2016rethinking}.
The result are displayed in Figure~\ref{fig:I1K}, middle line. These two tricks do not seem to have any effect on the training of the linear classifier.

In the bottom line of Figure~\ref{fig:I1K}, we implement a decrease of the learning rate with a  Cosine annealing (Cos) technique, in addition to (RA) and (LS). The (Cos) technique reduces the learning rate and enforces the {\em hyperconvergence mode}. As far as the accuracies are concerned, reducing the learning rate allows the algorithm to converge when the batches are small and is useless when the batch size is greater than 8K. This test corroborates the findings of~\cite{smith2017don} and the tests of Appendix~\ref{BS-CIFAR}.

In this benchmark, one of the important advantages of rescaling is to be able to perform several tests (batch reduction, repeated augmentation, label smoothing) without having to set the learning rate for each test.

\subsection{Influence of the averaging factor of the curvature}

\begin{figure}
    \centering
    \def\figwidth{.24\linewidth}
    \begin{subfigure}[b]{\figwidth}
        \centering
        \includegraphics[valign=m,width=\linewidth]{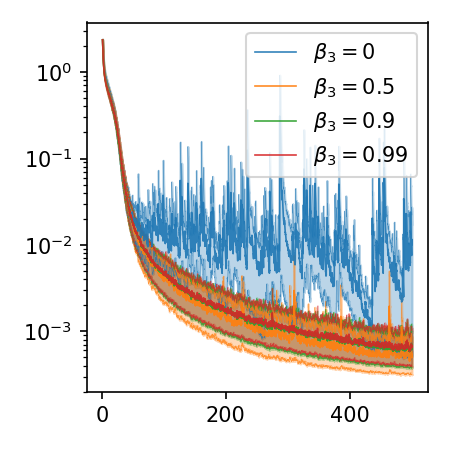}
        \caption{Training loss}
    \end{subfigure}~
    \begin{subfigure}[b]{\figwidth}
        \centering
        \includegraphics[valign=m,width=\linewidth]{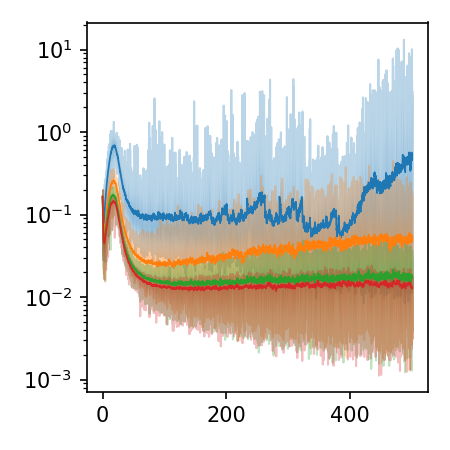}
        \caption{Step}
        \label{fig:impact_beta3:step}
    \end{subfigure}~
    \begin{subfigure}[b]{\figwidth}
        \centering
        \includegraphics[valign=m,width=\linewidth]{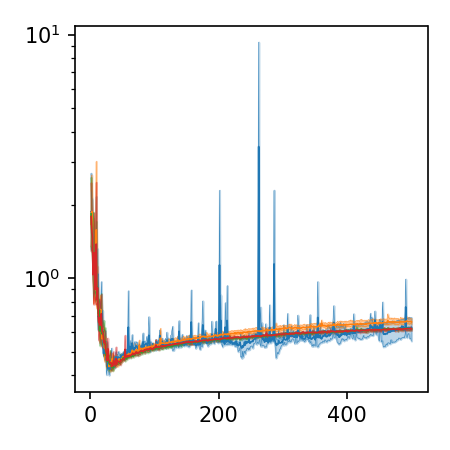}
        \caption{Testing loss}
    \end{subfigure}~
    \begin{subfigure}[b]{\figwidth}
        \centering
        \includegraphics[valign=m,width=\linewidth]{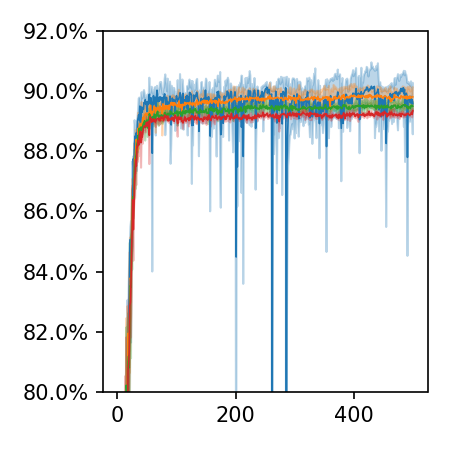}
        \caption{Accuracy test}
    \end{subfigure}
    \figspace\vspace{-0.2cm}
    \caption{Training loss, step size, testing loss and test accuracy on the CIFAR10 classifier with RED-SGD.
    The tests are conducted with different values of the curvature averaging parameter $\beta_3$.
    A value $\beta_3=0$ yields instability and $\beta_3\in[0.5, 0.99]$ has little impact on the convergence rate.
    }
    \label{fig:impact_beta3}
\end{figure}

This section is dedicated to the study of the impact of the averaging factor of the curvature $\beta_3$ on the algorithm.
A low value $\beta_3\simeq 0$ yields an estimation of the curvature that is less dependent of the past iterations at the expense of having a higher variance.
A value close to $1$ results in a low variance estimation but that has a bias due to old iterations.
In Figure~\ref{fig:impact_beta3}, the CIFAR10 classifier is optimized using RED-SGD with values of $\beta_3\in\{0, 0.5, 0.9, 0.99\}$.
Interestingly, the parameter that gives the fastest increase of the test accuracy is $\beta_3=0$ at the cost of more instabilities.
Although higher values of $\beta_3$ lead to an underestimation of the step size, the difference of performance on the training loss is insignificant.
Overall, a value $\beta_3\in[0.5, 0.99]$ has little impact on the convergence rate of the algorithm and a default value of $\beta_3=0.9$ can be considered.

\section{Conclusion and discussion}\label{sec:conclusion}

We developed a framework that allows automatic rescaling of the learning rate of a descent method with the use of the curvature, which is an easily affordable second order information computed by automatic differentiation. This rescaling yields a  data and direction adapted learning rate with a physical meaning. The practitioner can choose the behavior of the algorithm by setting the value of this rescaled learning rate. A value between $1/2$ and $1$ results in convergence, a value above $1$ yields hyperexploration of the space of parameters and a value below $1/2$ enforces convergence when stochasticity is of importance.

In the numerical examples of Section~\ref{sec:conv_expl_tradeoff} a choice of exponential decrease is competitive to simple manual tuning of the learning rate in the case of SGD and RMSProp preconditioning. 
In Section~\ref{sec:hyperexpl}, we show that the choice $\ell >1$ allows escaping basin of attraction of local minima.
The more intricated benchmark of Section~\ref{sec:hyperconv} show that rescaling doesn't save us from reducing the learning rate but that it allows to control the environment and compare different experiments.

The main limitation of this method is that it does not allow use of momentum. Indeed momentum methods do not necessarily yield directions of descent and do rely on per-iteration minimization of Lyapunov functions \cite{polyak2017lyapunov}. Implementing momentum methods with curvature computation is a challenge reserved for future works.
Another drawback is the need to use $\Cc^2$ activation functions, notably excluding ReLU.
Finally, the curvature computation, also affordable in theory, requires additional implementations on top of ready-to-use machine learning librairies, which restricts, for now, our method to rather simple networks.

\section*{Acknowledgement}
This work was supported by the ANR Micro-Blind.
F. de Gournay acknowledges the support of AI Interdisciplinary Institute ANITI funding, through the French ``Investing for the Future— PIA3'' program under the Grant Agreement ANR-19-PI3A-0004. This work was performed using HPC resources from GENCI-IDRIS (Grant 2021-AD011012210R1).

\clearpage
{\small
\bibliographystyle{plain}
\bibliography{biblio}
}

\clearpage
\appendix

\section{Second order computation}

\subsection{Hessian-vector dot product}\label{sec:Pearlmutter}

In this section, we turn our attention to showing how to compute $\nabla^2 \mathcal J(\Theta)\dot \Theta$ in our setting.
The results are known as the {\em Pearlmutter's trick} \cite{christianson1992automatic,pearlmutter1994fast}.
We emphasize that the computation of the curvature is simplier than the Hessian-vector product.
In our setting, the trick that allows the computation of the Hessian-vector product is based on the following ideas
\begin{itemize}
    \item The mapping $\dot \Theta \mapsto \f12 \langle\nabla^2 \mathcal J(\Theta)\dot \Theta,\dot \Theta\rangle$ is bilinear. If we differentiate with automatic differentiation this mapping with respect to $\dot \Theta$, we retrieve $\nabla^2 \mathcal J(\Theta)\dot \Theta$.
    \item Because $X(\Theta)$ is fixed, the aforementioned mapping is defined by a single forward recurrence. Hence, only one additional backward recurrence should be sufficient to compute $\nabla^2 \mathcal J(\Theta)\dot \Theta$.
\end{itemize} 

In order to make explicit this backward recurrence, we need to introduce two vectors $A_s\in \mathcal H_s$ and $B_s\in \mathcal G_s$ that are defined by the implicit equation:
\[\langle A_s,a\rangle_{\mathcal H_s}+\langle B_s,b\rangle_{\mathcal G_s}=\langle \nabla^2 \mathcal F_s (\dot x_s,\dot \theta_s)\otimes (a,b), \hat x_{s+1}\rangle_{\mathcal H_{s+1}} \quad \forall (a,b) \in \mathcal H_s\times \mathcal G_s.\]
The existence and uniqueness of $(A_s,B_s)$ is just Riesz theorem applied to the linear form on $\mathcal H_s\times \mathcal G_s$:
\[(a,b)\mapsto \langle \nabla^2 \mathcal F_s (\dot x_s,\dot \theta_s)\otimes (a,b), \hat x_{s+1}\rangle_{\mathcal H_{s+1}}.\]
Construct $\tilde X=(\tilde x_s)_s$ and $\tilde \Theta=(\tilde \theta_s)_s$ by a backward recurrence using
\begin{equation}
\label{eq:tildevariables}
    \begin{cases} 
\tilde x_s=(\partial_x \mathcal F)^* \tilde x_{s+1}+A_s\quad \text{ with }\tilde x_{n+1}=0 \\
\tilde \theta_s=(\partial_\theta \mathcal F)^* \tilde x_{s+1}+B_s.
    \end{cases}
\end{equation}
Then we have
\begin{equation}
\label{eq:Pearlmutter}
\nabla^2 \mathcal J(\Theta)\dot \Theta =\tilde \Theta
\end{equation}
In order to prove~\eqref{eq:Pearlmutter}, we show that for any other direction $\dot \Theta'$, 
we have
\[\langle \nabla^2 \mathcal J(\Theta)\dot \Theta,\dot \Theta'\rangle=
\langle \tilde \Theta,\dot \Theta'\rangle
 \]
First consider $\dot X'$ the tangent associated with direction $\dot \Theta'$. We have by bilinearity of $\nabla^2 \mathcal F_s$ and by Algorithm~\ref{alg:backpropwithcurv} that

\begin{equation}
\label{eqref:secondordersimplify}
\langle \nabla^2 \mathcal J(\Theta)\dot \Theta,\dot \Theta'\rangle=
\sum_s \langle \nabla^2 \mathcal F_s(\dot x_s,\dot \theta_s)\otimes (\dot x'_s,\dot \theta'_s), \hat x_{s+1}\rangle_{\mathcal H_{s+1}}=\sum_s \langle A_s,\dot x'_s\rangle+ \langle B_s,\dot \theta'_s\rangle
\end{equation}
By definition of $\tilde X$ and $\tilde \Theta$ in \eqref{eq:tildevariables} and by formula~\eqref{eq:directional:derivative_proof} for the tangent $\dot X'$, the following equality holds:
\begin{align*}
    &\langle \tilde x_s,\dot x'_s\rangle+\langle \tilde \theta_s,\dot \theta'_s\rangle -\langle A_s,\dot x'_s\rangle
    -\langle B_s,\dot \theta'_s\rangle\\
    &\quad =\langle(\partial_x \mathcal F)^* \tilde x_{s+1},\dot x'_{s}\rangle
    +\langle (\partial_\theta \mathcal F)^* \tilde x_{s+1},\dot \theta'_{s}\rangle=\langle \tilde x_{s+1},\dot x'_{s+1}\rangle
\end{align*}
Summing up the above equation for every $s$, using $\tilde x_{n+1}=0$, $\dot x'_0=0$ and \eqref{eqref:secondordersimplify} yields \eqref{eq:Pearlmutter}

\subsection{Structure of the layers}\label{sec:structlayers}

In this section, we explain how to compute the curvature for some of the standard layers used in DNNs.
First, we make clear the different kind of layers we use:
\begin{itemize}
\item {\bf Loss layers} are parameter-free layers from $\mathcal H_n$ to $\mathbb{R}$, they are denoted by $\mathcal L$
\[\Fc_n(x,\theta)=\mathcal L(x).\]
\item {\bf Smooth activation layers} do not have parameters and are such that $\mathcal H_{s+1}=\mathcal H_s$.
They are defined coordinate-wise through a smooth function $\Phi_s:\mathbb{R}\rightarrow\mathbb{R}$ with
\[\Fc_s(x,\theta)[i]=\Phi_s(x[i]) \quad \forall i.\]
\item {\bf Linear layers or convolutional layers}. The set of parameters are the weights (or kernel) denoted $\theta$. We suppose that these layers have no bias. They are abstractly defined as
\[\Fc_s(x,\theta)[i]=\sum_{k,j} \theta[k] x[j] \ind_{ijk} \]
where $i$ (resp $j,k$) denotes the sets of indices of the outputs (resp. the input, the weights). The function $(i,j,k)\mapsto \ind_{ijk}$ represents the assignment of the multi-index $(k,j)$ to $i$. This affectation is either equal to $1$ or $0$, that is $(\ind_{ijk})^2=\ind_{ijk}$. 
\item{\bf Bias layers} are layers where $\mathcal H_s=\mathcal H_{s+1}$ and are defined by
\[\Fc_s(x,\theta)[i]=x_s[i]+\sum_{k}\ind_{ik}\theta[k].\]
They are often concatenated with linear or convolutional layers. There is no restriction to split a biased linear layer into the composition of a linear layer and a bias layer. 
\item {\bf Batch normalization layers.} We split a batch normalization layer into the composition of four different layers, the centering layer, the normalizing layer, a linear layer with diagonal weight matrix and a bias layer. For each output index $i$, the centering and normalizing layers are defined by an expectation over the batch and some input indices. This expectation is denoted as $\E_i$. The centering layer can be written as
\[
\mathcal F_s(x,\theta)[i]=x_s[i]-\E_i(x_s).\]
The normalizing layer is defined as
\[\mathcal F_{s}(x,\theta)[i]=\frac{x[i]}{\sqrt{\E_i(x^2)+\varepsilon}}
.\]
\end{itemize}

For the different layers, we give the formula for the different recurrences in Table~\ref{tab:structlayers}.
We begin with the classic backward computations, they are mainly given here to settle the notations.

\begin{table}
    \centering
    \small
    \begin{tabular}{c|ccc}
    \toprule
    Name & $x_{s+1}[i]$ & $\hat x_{s}[j]$ & $\hat \theta_s[k]$ \\
    \midrule
    Activation & $\Phi(x_s[i])$ & $\Phi'(x_s[j])\hat x_{s+1}[j]$ & N.A. \\
    Linear & $\sum_{k,j} \theta[k] x_s[j] \ind_{ijk}$ & $\sum_{k,i} \theta[k] \hat x_{s+1}[i] \ind_{ijk}$ & $\sum_{j,i} \hat x_{s+1}[j] x_s[i]\ind_{ijk}$
    \\
    Bias & $x_s[i]+\sum_k \ind_{ik}\theta[k]$ & $\hat x_{s+1}[j]$ & $\sum_i \ind_{ik}\hat x_{s+1}[i]$ \\
    Centering & $x_s[i]-\E_i(x_s)$ & $\hat x_{s+1}[j]-\E_j(\hat x_{s+1})$ & N.A. \\
    Normalizing &
    $\begin{cases}\gamma=(\E_i(x_s^2)+\varepsilon)^{-1/2} \\
    x_{s+1}[i]=\gamma x_s[i] \end{cases}$
    & $\gamma \hat x_{s+1}[j]-x_s\E_j[\gamma^3x_s\hat x_{s+1}]$ & N.A. \\
    \bottomrule
    \end{tabular}

    \vspace{0.1cm}
    \begin{tabular}{c|cc}
    \toprule
    Name & $\dot x_{s+1}[i]$ & $r_s=\frac 1 2\langle \nabla^2 \mathcal F_s (\dot x_s,\dot \theta_s)\otimes (\dot x_s,\dot \theta_s),\hat x_{s+1}\rangle_{\mathcal H_{s+1}}$\\
    \midrule
    Activation & $\Phi'(x_s[i])\dot x_s[i]$ & $\frac 1 2\sum_i \Phi''(x_s[i])\dot x^2_s[i]\hat x_{s+1}[i]$\\
    Linear & $\sum_{k,j} \left(\theta[k] \dot x_s[j]+\dot\theta[k] x_s[j] \right)\ind_{ijk}$ & $\sum_{k,j,i} \dot \theta[k] \dot x_s[i]\hat x_{s+1}[j]\ind_{ijk}$
    \\
    Bias & $\dot x_s[i]+\sum_k \ind_{ik} \dot \theta[k]$ & $0$\\
    Centering & $\dot x_s[i]-\E_i(\dot x_s)$ & $0$\\
    Normalizing &
    $\begin{cases}\dot \gamma=-\E_i[\dot x_{s} x_s]\gamma^3 \\
    \dot x_{s+1}[i]=\gamma \dot x_s[i]+\dot \gamma x_s[i]
    \end{cases}$
    & $\begin{cases}\ddot \gamma=-\E_i(\dot x_s^2)\gamma^3+3\E_i(\dot x_s x_s)\gamma^5 \\
    r_s=\sum_i \frac 1 2 \left(\dot \gamma \dot x_s[i] +\ddot \gamma x_s[i]\right)\hat x_{s+1}[i]\end{cases}$ \\
    \bottomrule
    \end{tabular}
    \caption{Quantities needed in the forward, backward and second order passes for standard layers.}
    \label{tab:structlayers}
\end{table}

\section{Description of the numerical experiments}\label{sec:xpdescription}

All the experiments were conducted and timed using Python 3.8.11 and PyTorch 1.9 on an Intel(R) Xeon(R) W-2275 CPU @ 3.30GHz with an NVIDIA Quadro RTX 5000 GPU.
We also used the Jean-Zay HPC facility for additional runs.

The models are trained with a batch size of $256$ so that one epoch corresponds to $235$ iterations for MNIST and $196$ for CIFAR.
The number of epochs is set to $200$ for the MNIST classification and $500$ for the others.
Table~\ref{tab:summarydatasets} summarizes the characteristics of the datasets used.

Concerning the tuning of the standard methods, the step size and its decay factor were searched on a grid for the SGD and RMSProp methods.
The learning rate is constant per epoch and its value at the $n$\textsuperscript{th} epoch is given by
\[\tau_n=\tau_0d^n.\]
We searched amongst the values $\tau_0\in\{ 1\times 10^{n}, 5\times 10^{n}\}_{-5\leq n\leq 1}$ for the step size and $d\in\{0.97, 0.98, 0.99, 1\}$ for the step decay on MNIST classification and $d\in\{0.99, 0.995, 1\}$ for the others.
After $20\%$ of the total number of epochs, the couple $(\tau_0,d)$ that achieves the best training loss decrease is chosen.

For the CIFAR experiments we used data augmentation with a random crop and an horizontal flip.
In the CIFAR100 training we added a random rotation of at most $\pm 15$\textdegree.

For reproductibility, the values used in the experiments are summarized in Table~\ref{tab:summarydefaultxp}.
Unless explicitely stated, these are the default values used in the experiments of this work.
The computing time per epoch is reported in Table~\ref{tab:summarydefaultxp} for each method.
The codes of RED are not optimized, especially for the convolution layers where the backward with respect to the parameters is implemented by an additional run of the forward.
This explains why RED is twice slower than the standard methods on the CIFAR classifiers which make intensive use of convolution layers.

\begin{table}
    \centering
    \begin{tabular}{|l|c|c|c|c|}
        \toprule
        Dataset & MNIST & CIFAR10 & CIFAR100 \\
        \midrule
        License & CC BY-SA 3.0 & MIT License & Unknown \\
        Size of the training set & $60000$ & $50000$ & $50000$ \\
        Size of the testing set & $10000$ & $10000$ & $10000$ \\
        Number of channels & $1$ & $3$ & $3$ \\
        Size of the images & $28\times 28$ & $32\times 32$ & $32\times 32$ \\
        Number of classes & 10 & 10 & 100 \\
        \bottomrule
    \end{tabular}
    \caption{Summary of the datasets used.}
    \label{tab:summarydatasets}
\end{table}

\begin{table}
    \centering
    \renewcommand{\arraystretch}{1.7}
    \renewcommand{\cellalign}{tl}
    \renewcommand{\theadalign}{tl}
    \def\widthcol{0.12\textwidth}
    \def\widthcolleft{0.25\textwidth}
    \begin{tabular}{|l|c|c|c|c|}
        \toprule
        Type of problem & \makecell[cm{\widthcol}]{\centering MNIST\\ classification} & \makecell[cm{\widthcol}]{\centering CIFAR10\\ classification} & \makecell[cm{\widthcol}]{\centering CIFAR100\\ classification} & \makecell[cm{\widthcol}]{\centering MNIST\\ autoencoder} \\
        \midrule
        Type of network & \makecell[cm{\widthcol}]{\centering LeNet\\Dense} & \makecell[cm{\widthcol}]{\centering VGG11\\Convolutional} & \makecell[cm{\widthcol}]{\centering VGG19\\Convolutional} & {Dense} \\
        Activation functions & Tanh & SoftPlus $\beta=5$ & SoftPlus $\beta=5$ & ELU \\
        \red{$L^2$ regularization} & $\lambda=10^{-7}$ & $\lambda=10^{-7}$ & $\lambda=10^{-7}$ & $\lambda=10^{-7}$ \\
        Loss function & Cross entropy & Cross entropy & Cross entropy & MSE \\
        Number of epoch & 200 & 500 & 500 & 500 \\
        Batch size & 256 & 256 & 256 & 256 \\
        \makecell[cm{\widthcolleft}]{Number of epoch\\ for tuning} & 40 & 100 & 100 & 100 \\
        Iteration per epoch & 196 & 235 & 235 & 196 \\
        \makecell[cm{\widthcolleft}]{Computing time per epoch with the standard SGD / RMSProp} & $5.3$s / $5.4$s & $16.4$s / $16.8$s & $36.3$s / $36.9$s & $5.1$s / $5.3$s \\
        \makecell[cm{\widthcolleft}]{Computing time per epoch with RED-SGD / RED-RMSProp} & $7.6$s / $7.7$s & $30.1$s / $30.5$s & $65$s / $65$s & $5.7$s / $5.9$s \\
        \bottomrule
    \end{tabular}
    \caption{Summary of the experiment parameters.}
    \label{tab:summarydefaultxp}
\end{table}

\section{Additional numerical experiments}\label{sec:additionalxp}

\subsection{Dealing with momentum}\label{sec:momentum}

In Section~\ref{sec:conv_expl_tradeoff}, only stochastic optimizers without momentum are presented. In this section, we discuss the extension of our algorithm to momentum based update directions, notably momentum with RMSProp preconditioning which is the celebrated Adam algorithm~\cite{kingma2014adam}.

Incorporating momentum consists in replacing the gradient by an exponential moving average of the past iterates of the gradients with a parameter $\beta_1 \in [0,1[$.
In our setting, it amounts to replacing line~\ref{lst:line:dir_descent} of Algorithm~\ref{alg:RED} by lines~\ref{lst:algmomentum:ema_grad} and \ref{lst:algmomentum:update_dir} of Algorithm~\ref{alg:modif_momentum}.
\begin{algorithm}
    \caption{Adding momentum to RED}
    \begin{algorithmic}[1]
        \State {\bf Initialization} $\hat g_0=0$.
        \For{$k=1..$}
            \State $\cdots$
            \State $\hat g_k=\beta_1 \hat g_{k-1}+(1-\beta_1)g_{k}\quad$ and $\quad\tilde g_k=\hat g_{k}/(1-\beta_1^k)$ \label{lst:algmomentum:ema_grad}
            \State $\dot \Theta_k=P_k^{-1}\tilde g_k$ \label{lst:algmomentum:update_dir}
            \State $\cdots$
        \EndFor
    \end{algorithmic}
    \label{alg:modif_momentum}
\end{algorithm}

Momentum was introduced by Polyak \cite{polyak1964some} in the convex non-stochastic setting.
It can be interpreted as an adaptation of a convex method to a non-convex stochastic problem. We coin this explanation as the {\em heavy-ball} analysis.
Another point of view, which we denote as {\em variance reduction}, is that the exponential moving average $\tilde g_k$ is a better estimator of $\nabla \Jc(\Theta_k)$ than $g_k$. Indeed all the previous batches $(\Bc_m)_{m\le k}$ are taken into account in the computation of $\tilde g_k$. The downside is that the averaged quantity is $\nabla\Jc_{\Bc_m}(\Theta_m)$ and not $\nabla\Jc_{\Bc_m}(\Theta_k)$, this introduces a bias in the estimation of $\nabla\Jc(\Theta_k)$. With this interpretation in mind, the parameter $\beta_1$ which drives the capacity of the exponential moving average to forget the previous iterations has to be tuned between the mini-batches gradient variance (high variance leads to high $\beta_1$) and the convergence (high values of $\Vert \Theta_k - \Theta_{k-1}\Vert$ lead to low choice of $\beta_1$). In \cite{kingma2014adam}, the authors propose to solve this dilemna by taking decaying values of $\beta_1$, although in practice, the parameter $\beta_1$ is constant.

\paragraph{Momentum: heavy ball or variance reduction?}
When momentum is understood as an heavy-ball method, at iteration $k$ there are no reasons for $-\dot\Theta_k$ to be a direction of descent.
Because our algorithm relies on the assumption that $-\dot\Theta_k$ is a direction of descent to choose a step, our analysis falls apart and RED should be used with care. On the other hand, if momentum is a variance reduction technique, the step has to be taken small enough in order not to bias the gradient estimation. With this latter assumption, RED can be applied.

\begin{figure}
    \centering
    \def\figwidth{.24\linewidth}
    \begin{tabular}{@{}c@{}c@{}c@{}c@{}c@{}}
        \rotatebox[origin=c]{90}{\small $\ell=1$} &
        \includegraphics[valign=m,width=\figwidth]{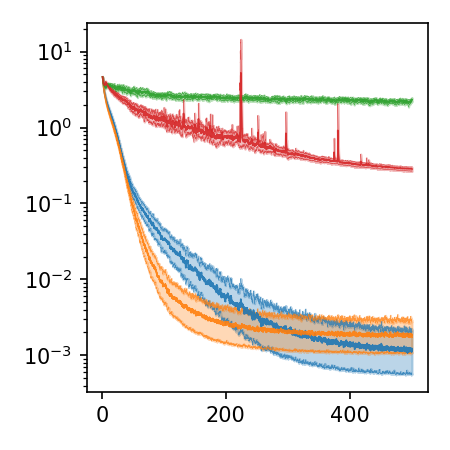} &
        \includegraphics[valign=m,width=\figwidth]{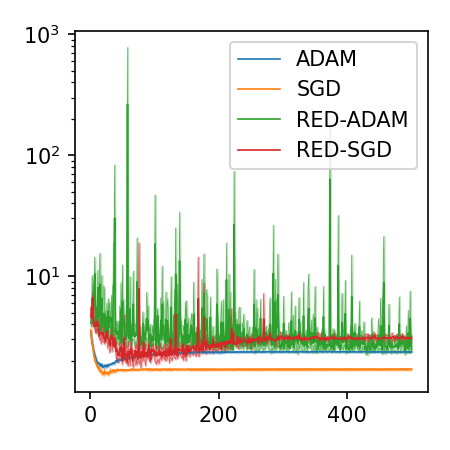} &
        \includegraphics[valign=m,width=\figwidth]{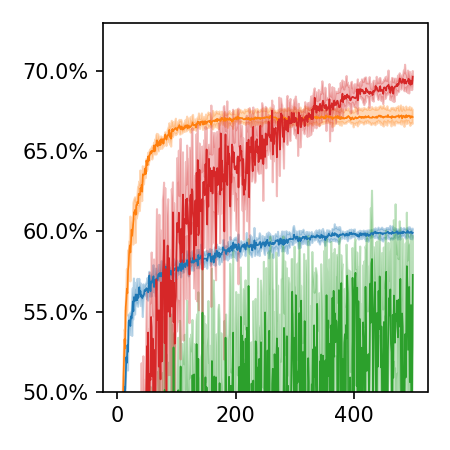} &
        \includegraphics[valign=m,width=\figwidth]{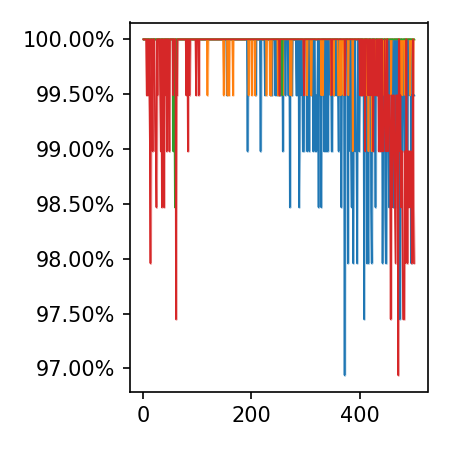} \\

        \rotatebox[origin=c]{90}{\small $\ell=0.1$} &
        \includegraphics[valign=m,width=\figwidth]{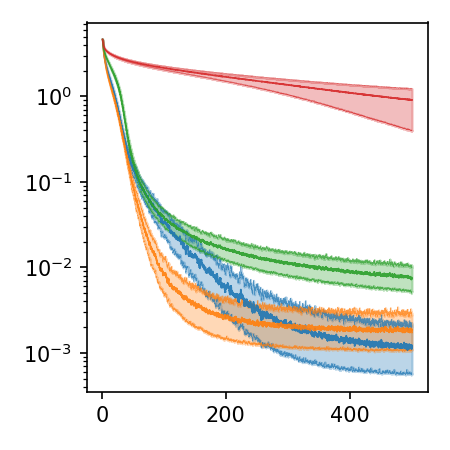} &
        \includegraphics[valign=m,width=\figwidth]{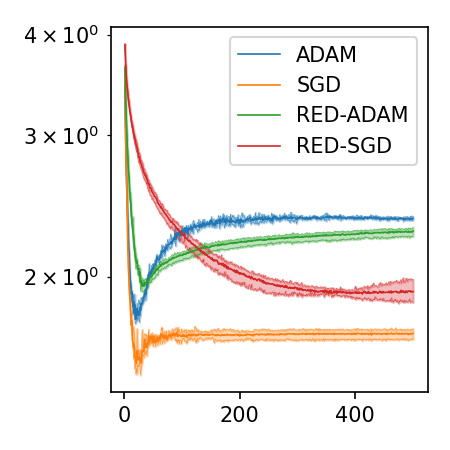} &
        \includegraphics[valign=m,width=\figwidth]{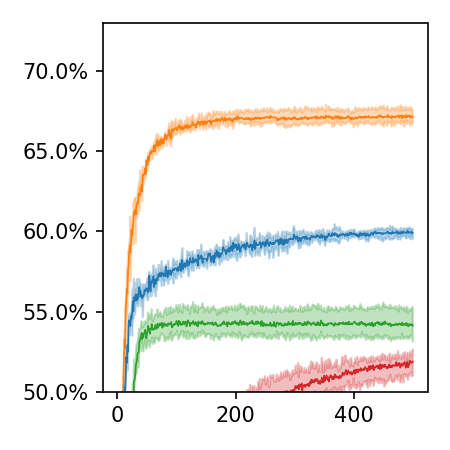} &
        \includegraphics[valign=m,width=\figwidth]{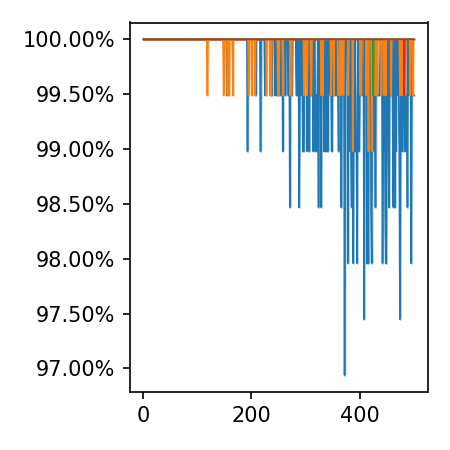} \\
        
        & Training loss & Testing loss & Accuracy test & Percent. dir. descent $q_n$
    \end{tabular}
    \caption{Tests with momentum ($\beta_1=0.9$) with and without the learning rate stabilization ($\ell=1$ or $\ell=0.1$) for RED on CIFAR100.
    Manually-tuned algorithms (orange for SGD, blue for Adam) and their RED version (red for SGD, green for Adam) are given.
    Lower learning rate in RED ensures that momentum yields direction of descent at the expense of loosing the exploration of the set of parameters.
    }
    \label{fig:results_momentum}
\end{figure}

In order to determine if, in our case, momentum acts as an heavy ball method or as a variance reduction technique, we study numerically when the standard Adam and SGD with momentum optimizers yield a direction of descent. 
In Figure~\ref{fig:results_momentum} first row, the test of CIFAR100 in Section~\ref{sec:autovsmanual} is performed with a momentum $\beta_1=0.9$ and the hyperparameters were tuned using the same policy (see Appendix~\ref{sec:xpdescription}).
We display in the last column of Figure~\ref{fig:results_momentum} first row, the percentage of direction of descent per epoch with respect to the current batch $\Bc_k$.
If $n$ is the epoch number and $\mathcal K_n$ the set of the iterations that are in epoch $n$, this percentage is given by:
\begin{equation}
    q_n = \frac{1}{|\mathcal K_n|}\sum_{k\in\mathcal K_n} \ind_{\langle g_k, \dot\Theta_k \rangle \geq 0}
\end{equation}

We observe that on classification problems, SGD with momentum and more particularly Adam yield directions of update that are not direction of descent for $\Jc_{\Bc_k}$.

\paragraph{Step choice}
The RED algorithm needs a rule to deal with update directions which are not directions of descent.
One possibility is to allow negative steps, which we discard since this would annihilate the heavy-ball effect. Another possibility, which we retain, is to take the absolute value of $\tau_k^\star$ in line~\ref{lst:line:step}. In a nutshell, compute the step for the opposite direction (which is a direction of descent) and use this step in the current direction. This choice is arbitrary and to properly tackle the momentum case, interpretations using Lyapunov functions should be considered. The choice of such functions is not clear and we defer such an analysis to future work.

In Figure~\ref{fig:results_momentum} first row the results of the optimization using RED on CIFAR100 with momentum are displayed.
The parameters for the initial learning rate and its decay factor are the default ones $\ell=1$ and $\eta=1/2$.
The RED algorithm has difficulties to converge both on the training and testing losses.
We observed that the steps chosen by RED are several orders of magnitude higher than the ones obtained by manual tuning.
On classification problems, RED follows directions of update that are not direction of descent.

\paragraph{Learning rate multiplication}
The impediment to using RED with momentum is that directions of update are not directions of descent. This can be solved by reducing the initial learning rate $\ell$ to take smaller steps $\tau_k$ so that $\Vert \Theta_k - \Theta_{k-1}\Vert$ remains small.

We propose to diminish the initial learning rate by using $\ell=1-\beta_1$.
This choice may seem arbitrary but it is inspired by the proofs of convergence of \cite{defossez2020simple} that have bounds which scale as $1-\beta_1$.
The experiments of Figure~\ref{fig:results_momentum} last row are performed with the same set of parameters except for the initial learning rate which is set to $\ell=0.1$.
With this smaller learning rate, the algorithm is stable and converges.
Of importance, RED always yield direction of descent as seen from the bottom-right of Figure~\ref{fig:results_momentum}.
As $\ell$ was decreased, the exploration is lost, explaining these poor convergence results.

\paragraph{Conclusion}
When using momentum, decreasing the learning rate makes the experiments fit in the framework the algorithm was proposed for.
As a consequence, this causes the loss of the exploration which is critical to speed-up the convergence.
The correct way of dealing with momentum would be to identify the Lyapunov function that has to be minimized, which is left for future work.

\subsection{Batch reduction on CIFAR}\label{BS-CIFAR}

\begin{figure}
    \centering
    \def\figwidth{.24\linewidth}
    \begin{tabular}{@{}c@{}c@{}c@{}c@{}c@{}}
        \rotatebox[origin=c]{90}{\small Training loss} &
        \includegraphics[valign=m,width=\figwidth]{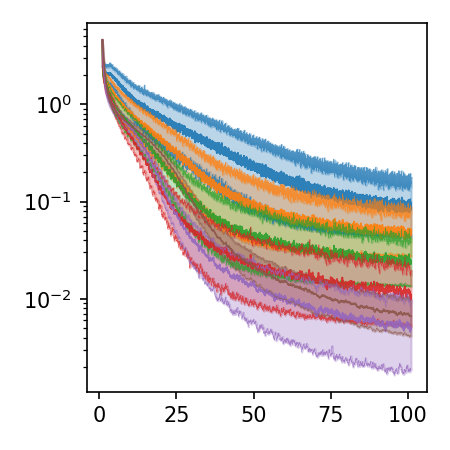}  &
        \includegraphics[valign=m,width=\figwidth]{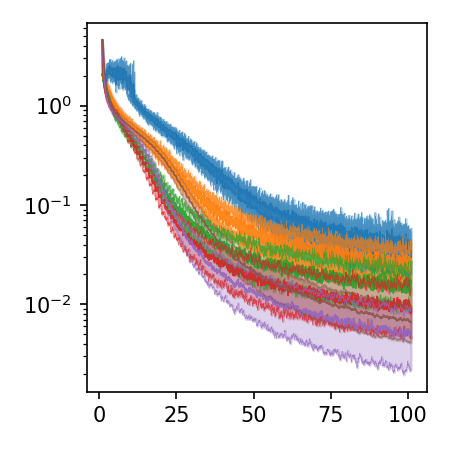}     &
        \includegraphics[valign=m,width=\figwidth]{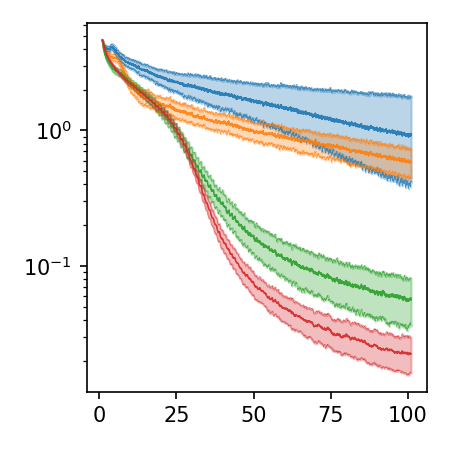} &
        \includegraphics[valign=m,width=\figwidth]{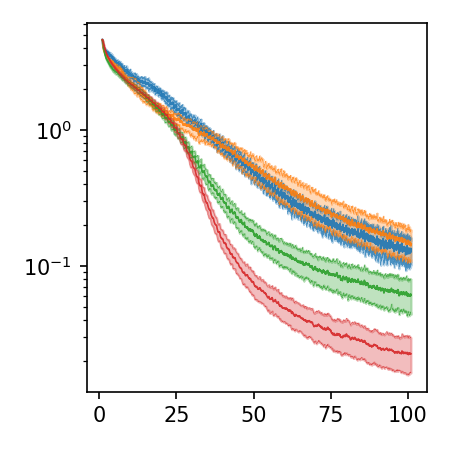}
        \\

        \rotatebox[origin=c]{90}{\small Testing loss} &
         \includegraphics[valign=m,width=\figwidth]{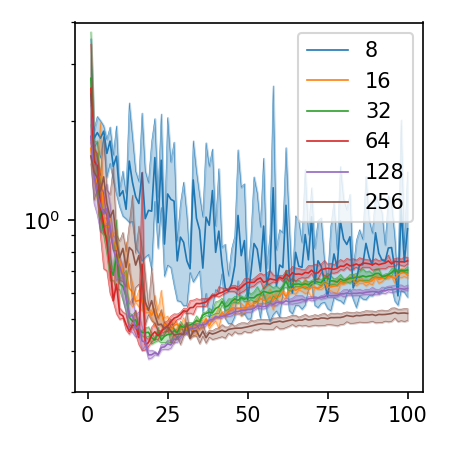}  &
        \includegraphics[valign=m,width=\figwidth]{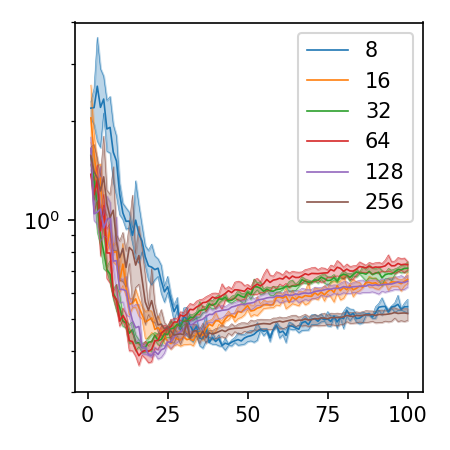}     &
        \includegraphics[valign=m,width=\figwidth]{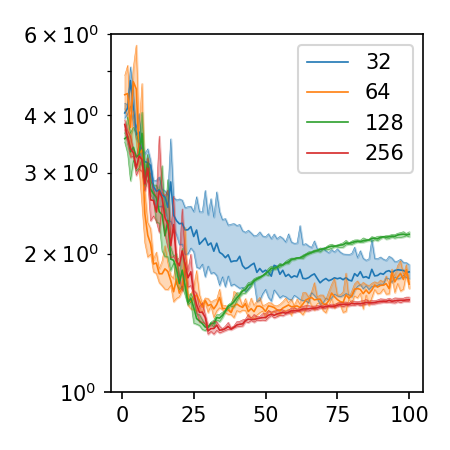} &
        \includegraphics[valign=m,width=\figwidth]{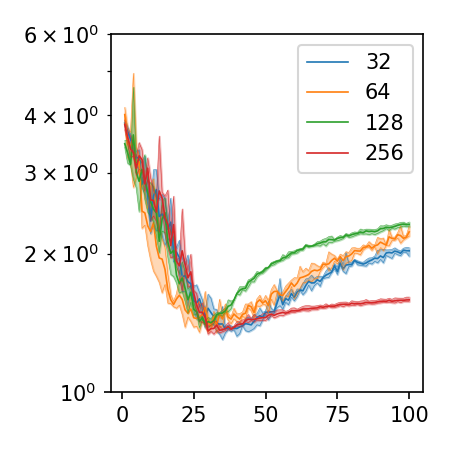}
        \\

        \rotatebox[origin=c]{90}{\small Accuracy test} &
        \includegraphics[valign=m,width=\figwidth]{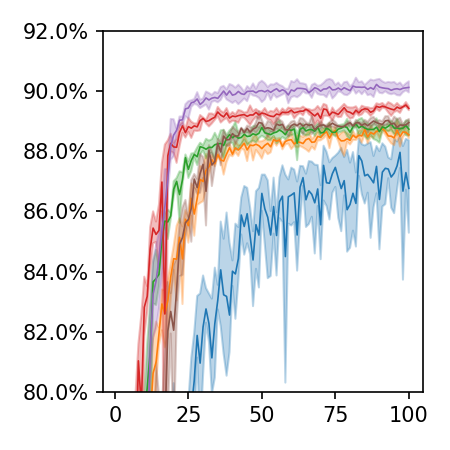}  &
        \includegraphics[valign=m,width=\figwidth]{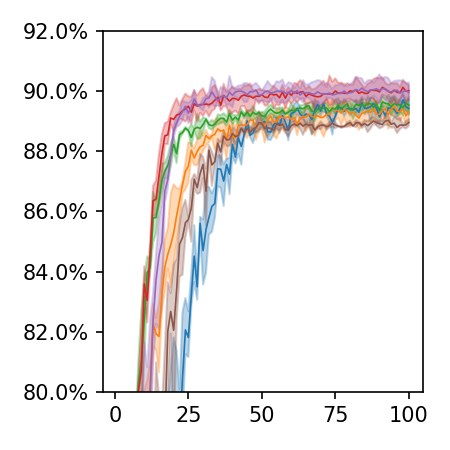}     &
        \includegraphics[valign=m,width=\figwidth]{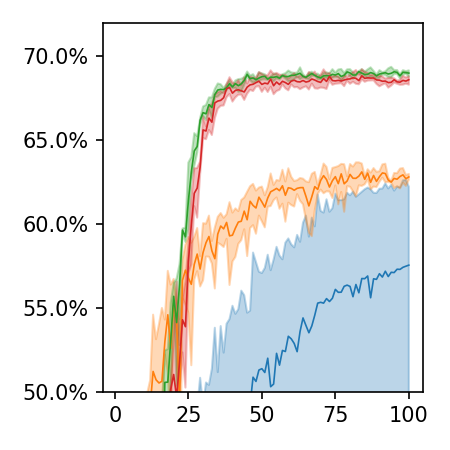} &
        \includegraphics[valign=m,width=\figwidth]{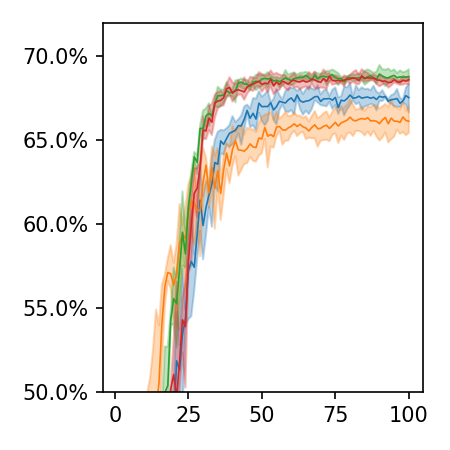}
        \\
        & CIFAR10 &
        mem-CIFAR10 &
        CIFAR100 &
        mem-CIFAR100
    \end{tabular}
    \figspace
    \caption{Batch reduction on the CIFAR10 and CIFAR100 datasets. The columns 1 and 3 are the vanilla RED-algorithm and the column 2 and 4 are the patches that (partially) solve the problem when the batch size is smaller than the number of classes. 
    }
    \label{fig:batch}
\end{figure}

In this section, we study batch dependence on the RED method for the CIFAR datasets. Reducing the batch size mimicks harder stochastic problems while keeping the experiment in a controlled environment. In Figure~\ref{fig:batch} (column 1 and 3), we provide the results obtained for different batch size and the evolution of the training and testing loss functions per epoch. The RED parameters are an initial learning rate $\ell=1$ and a target learning rate $\ell=\frac 1 2$ after $100$ epochs. A striking phenomenon in Figure~\ref{fig:batch} (column 1 and 3) is the loss of performance of the algorithm when the batch size is smaller than the number of classes. 

Because of the relationship between  the batch size and the number of classes, we wish to study if the last layer -- the Linear Classifier (LC) -- is the layer the most impacted by the batch size reduction. The LC optimizes the parameters of hyperplanes (one per class) which separate the information given by the remaining of the network, the Feature Extractor (FE)
When the batch size is too small, some classes are not represented in the batch. The corresponding hyperplanes receive update information which is oblivious to the data of their class. We believe that this effect explains the loss of performance of the LC and a lack of precision in the computation of the curvature.

In order to verify our assumption, we implement a {\em memory} layer, which is set between the FE and the LC. This memory layer stores the last 256 data given by the FE. We coin this trick a memory-DNN.
Because the LC is fed with this memory, it should behave as if the batch size was 256, although the memory suffers from a slight delay, due to the fact that it is not updated for the current parameters of the FE. The memory footprint and computational load of the memory-DNN is increased by a small factor, since the FE is fed with small batches and is responsible for most of the computational load and memory footprint.
In Figure~\ref{fig:batch} (column 2 and 4), we collect the results of memory-DNN.
Of striking importance is a better behavior of memory-DNN compared to the standard DNN when the batch size is smaller than the number of classes.

In this test, we provide a simple remedy to avoid stochasticity issues in the training of the LC in a classification problem. More important than the memory trick is the fact that rescaling the learning rate allows us to provide a unified environment to test the method. The learning rate do not have to be adapted for each experiment, which would eventually prevent us from drawing any conclusions.

\subsection{Effect of the $L^2$ regularization}\label{sec:regularization_effect}

\begin{figure}
    \centering
    \def\figwidth{.24\linewidth}
    \begin{tabular}{@{}c@{}c@{}c@{}c@{}c@{}}
        \rotatebox[origin=c]{90}{\small CIFAR10 $\lambda=10^{-7}$} &
        \includegraphics[valign=m,width=\figwidth]{cifar10_loss-train.png} &
        \includegraphics[valign=m,width=\figwidth]{cifar10_step.png} &
        \includegraphics[valign=m,width=\figwidth]{cifar10_loss-test.png} &
        \includegraphics[valign=m,width=\figwidth]{cifar10_acc-test.png} \\

        \rotatebox[origin=c]{90}{\small CIFAR10 $\lambda=10^{-4}$} &
        \includegraphics[valign=m,width=\figwidth]{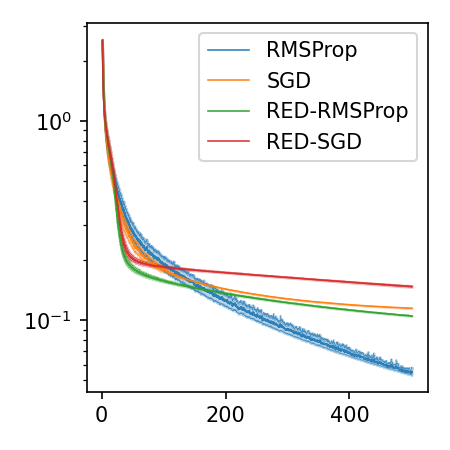} &
        \includegraphics[valign=m,width=\figwidth]{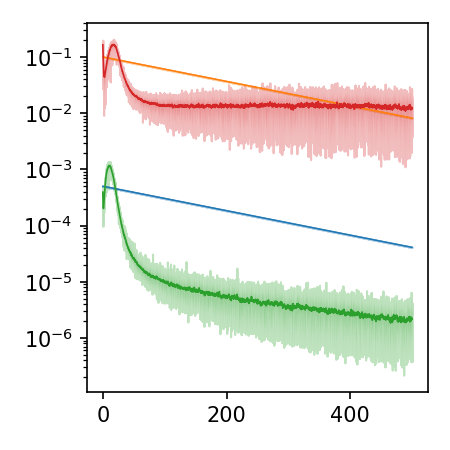} &
        \includegraphics[valign=m,width=\figwidth]{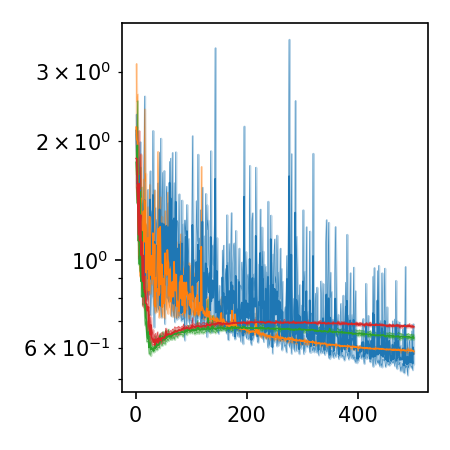} &
        \includegraphics[valign=m,width=\figwidth]{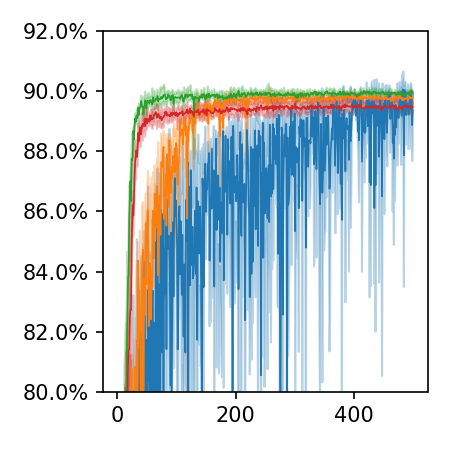} \\

        & Training loss & Step & Testing loss & Accuracy test
    \end{tabular}
    \caption{Influence of the \red{$L^2$ regularization} $\lambda$.
    Manually-tuned algorithms (orange for SGD, blue for RMSProp) and their RED version (red for SGD, green for RMSProp) are given.
    For these tests, the smaller the \red{regularization}, the best the convergence of RED.
    }
    \label{fig:results_weightdecay}
\end{figure}

According to the paragraph on the $L^2$ regularization in Section~\ref{sec:analysis_rescaling}, a regularization is introduced in our algorithm to counteract the effet of a potentially vanishing Hessian in the direction of update.
This is a theoretical limitation and we study in this section the influence of \red{this regularization} on the performance of RED.
We conduct the experiments of Section~\ref{sec:autovsmanual} for the CIFAR10 dataset with different \red{values of the regularization} $\lambda\in\{10^{-7},10^{-4}\}$.
The hyperparameters of the standard SGD and RMSProp optimizers are tuned for each value of $\lambda$.
We report in Figure~\ref{fig:results_weightdecay} the different results, including the ones that are shown in Section~\ref{sec:autovsmanual}.
The grid search on the training loss that led to the choice of parameters for RMSProp and $\lambda=10^{-4}$ yielded large step size at the cost of instabilities in the test metrics.
On all test cases, we observe that a value of \red{regularization} close to zero ($\lambda=10^{-7}$) gives good convergence results.
In the considered tests, the need of a \red{regularization} seems to be more of a theoretical limitation than a practical one.

\subsection{Comparison with BB and Robbins-Monro}\label{sec:comparisonexistingalgo_appendix}

In this section we compare our algorithm with the closest existing approach \cite{castera2022second}, named \emph{step-tuned}, where the authors approximate the curvature with a BB method. \red{We also compare it with a Robbins-Monro decay rule of the learning rate for SGD.}
We did not compare with the BB method of \cite{yang2018random} as this method requires the computation of the gradient over the whole dataset at each epoch.

The step-tuned optimizer has several hyperparameters and we use the default ones except the learning rate as advised in \cite{castera2022second}.
The initial learning rate of \cite{castera2022second} is searched on the same grid than the standard SGD (see Appendix~\ref{sec:xpdescription}).

In Figure~\ref{fig:compsteptuned}, the results of the optimization on the CIFAR10 classifier and on the autoencoder are given for two values of the \red{$L^2$ regularization} $\lambda\in\{10^{-7},10^{-4}\}$.
RED algorithm is outperformed by step-tuned only on the training loss but the learning rate of step-tuned has been optimized for the training loss and we cannot expect better performance than step-tuned on this criterion.
Finally, RED is more stable on every test metrics and has better generalization than step-tuned.
Step-tuned \cite{castera2022second} requires the optimization of the learning rate and because RED does not need any hyperparameter adjustment, our method is competitive with this existing work. Note also that step-tuned is not available with the RMSProp preconditioner, when RED handles any kind of preconditioning technique.

\begin{figure}
    \centering
    \def\figwidth{.24\linewidth}
    \begin{tabular}{@{}c@{}c@{}c@{}c@{}c@{}}
        \rotatebox[origin=c]{90}{\small CIFAR10 $\lambda=10^{-7}$} &
        \includegraphics[valign=m,width=\figwidth]{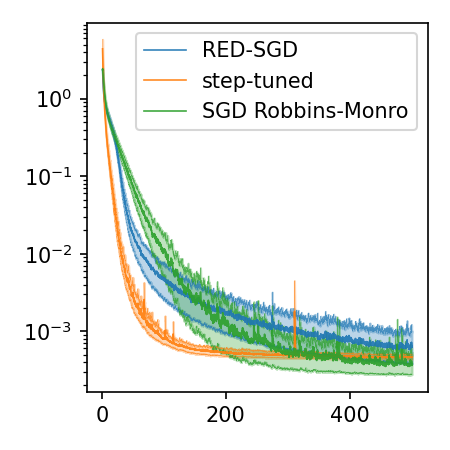} &
        \includegraphics[valign=m,width=\figwidth]{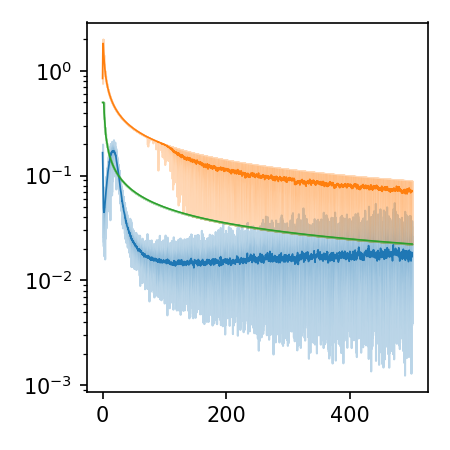} &
        \includegraphics[valign=m,width=\figwidth]{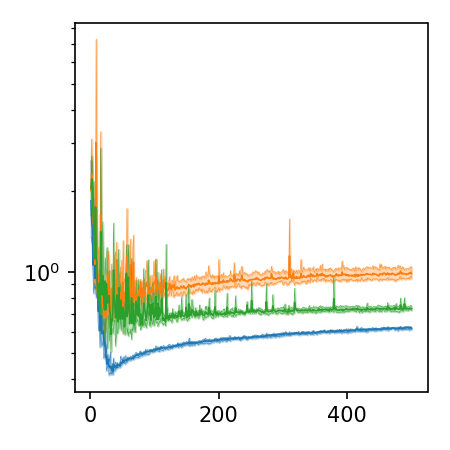} &
        \includegraphics[valign=m,width=\figwidth]{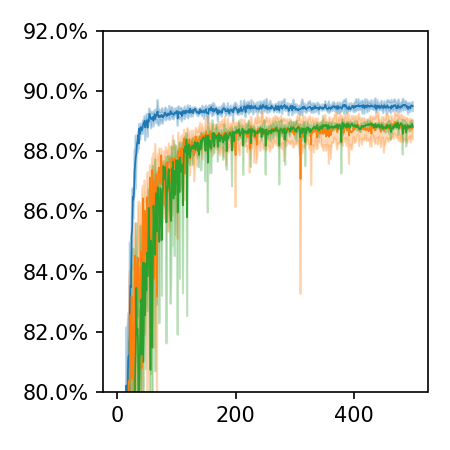} \\

        \rotatebox[origin=c]{90}{\small CIFAR10 $\lambda=10^{-4}$} &
        \includegraphics[valign=m,width=\figwidth]{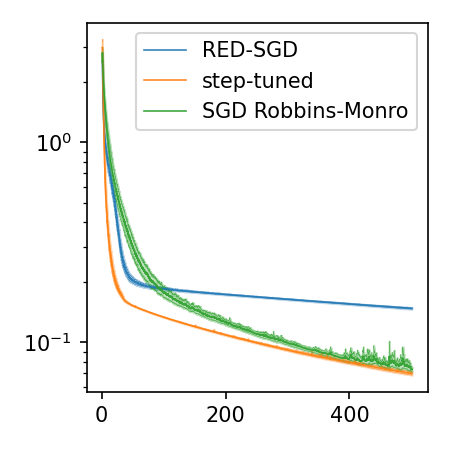} &
        \includegraphics[valign=m,width=\figwidth]{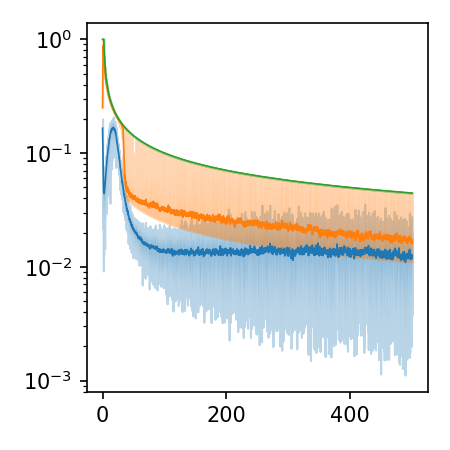} &
        \includegraphics[valign=m,width=\figwidth]{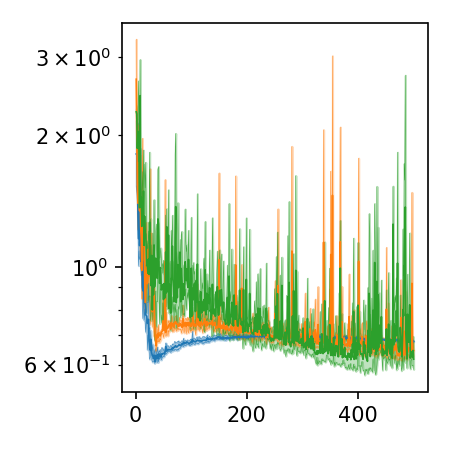} &
        \includegraphics[valign=m,width=\figwidth]{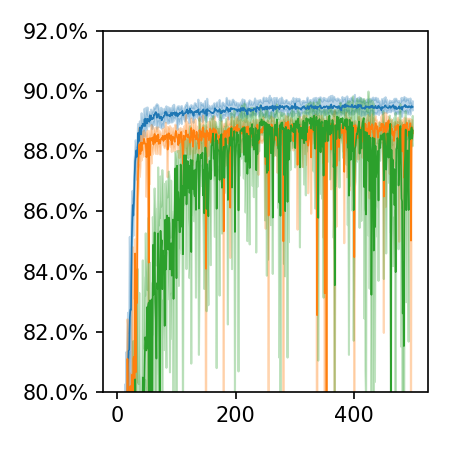} \\

        & Training loss & Step & Testing loss & Accuracy test
    \end{tabular}
    \caption{Comparison with step-tuned \cite{castera2022second} method on the CIFAR10 classifier.
    Standard SGD (blue), RED (orange) and step-tuned (green) are given.
    RED is competitive with step-tuned on the accuracy but not on the training loss of the CIFAR10 classifier for which step-tuned is optimized.
    }
    \label{fig:compsteptuned}
\end{figure}

\end{document}